\pdfoutput=1
\documentclass{article}
\usepackage{layout}
\usepackage{theorem}
\usepackage{amsmath}
\usepackage{amssymb}
\usepackage{epsfig,graphics,color}
\usepackage{graphicx}
\usepackage{enumerate}
\usepackage{float}
\usepackage{hhline}
\usepackage{version}
\usepackage{fancyheadings}
\usepackage{rotating}
\RequirePackage[colorlinks,citecolor=blue,urlcolor=blue]{hyperref}
\usepackage[ruled]{algorithm2e}
\usepackage{subcaption}
\usepackage{gensymb}

\newtheorem{lemma}{Lemma}[section] %%    with section number.

\newtheorem{prop}{Proposition}[section]
\newtheorem{definition}{Definition}[section]
\newtheorem{condition}{Condition}[section]
\newcommand{\bed}{\begin{definition}}
\newcommand{\eed}{\end{definition}}
\newcommand{\beq}{\begin{equation}}
\newcommand{\eeq}{\end{equation}}

\newcommand{\bitem}{\begin{itemize}}
\newcommand{\eitem}{\end{itemize}}

\newcommand{\margmin}{\mathrm{argmin}}

\newcommand{\beqn}{\begin{equation}}
\newcommand{\eeqn}{\end{equation}}
\newcommand{\balign}{\begin{align}}
\newcommand{\ealign}{\end{align}}

\usepackage{amssymb}

\DeclareFontFamily{OT1}{pzc}{}
\DeclareFontShape{OT1}{pzc}{m}{it}{<-> s * [1.10] pzcmi7t}{}
\DeclareMathAlphabet{\mathpzc}{OT1}{pzc}{m}{it}

\newtheorem{thm}{\sc Theorem}

\begin{document}

\title{Principal Boundary on Riemannian Manifolds}

\maketitle

\author{\noindent Zhigang Yao\\
Department of Statistics and Applied Probability  \\ 21 Lower Kent Ridge Road\\ 
National University of Singapore, Singapore 117546  \\ 
\noindent  email: \texttt{zhigang.yao@nus.edu.sg}\\  

\noindent  Zhenyue Zhang\\
Department of Mathematics\\
Zhejiang University, Yuquan Campus\\ Hangzhou 310027 China\\ 
\noindent  email: \texttt{zyzhang@math.zju.edu.cn}}

\vspace{0.2 in}

\begin{abstract}
We consider the classification problem and focus on nonlinear methods for classification on manifolds. For multivariate datasets lying on an embedded nonlinear Riemannian manifold within the higher-dimensional ambient space, we aim to acquire a classification boundary for the classes with labels, using the intrinsic metric on the manifolds. Motivated by finding an optimal boundary between the two classes, we invent a novel approach -- the principal boundary. From the perspective of classification, the principal boundary is defined as an optimal curve that moves in between the principal flows traced out from two classes of data, and at any point on the boundary, it maximizes the margin between the two classes. We estimate the boundary in quality with its direction, supervised by the two principal flows. We show that the principal boundary yields the usual decision boundary found by the support vector machine in the sense that locally, the two boundaries coincide. Some optimality and convergence properties of the random principal boundary and its population counterpart are also shown. We illustrate how to find, use and interpret the principal boundary with an application in real data.
\end{abstract}

\vspace*{.3in}


%\newpage

\section{Introduction}
\label{sec:Intro}

Most of the classification methodology in high dimensional data analysis is deeply rooted in methods relying on linearity. Modern data sets often consist of a large number of samples, each of which is made up of many more features. Manifold data arises in the sense that the sample space of data is fundamentally nonlinear. Rather than viewing any observation as a point in a high-dimensional Euclidean space, it is more natural to assume the data points lie on an embedded lower-dimensional non-linear manifold within the higher-dimensional space. The lower-dimensional manifold structure can usually be interpreted from at least two scenarios: 1) the physical data space is an actual manifold; 2) the underlying data structure can be approximated by a close manifold. In the former scenario, the data space is usually known, and it can be further seen as data in the shape space \cite{Kendall1999, Patrangenaru2015}, e.g., a seismic event in geophysics, leaf growth pattern, and data with non-linear constraints. It is thus forced to lie on a manifold. In the latter scenario, the manifold is uncovered from the data set by a non-linear dimensionality reduction technique referred to as the manifold learning method \cite{Roweis2000, Donoho2003, zhang2004}, and is thus considered unknown.

In this work, we consider the classification problem and design the non-linear methods that perform as a boundary for classifying data sets lying on the {\it known} manifold. This problem has become increasingly relevant, as many real applications such as medical imaging \cite{Gerber2010, Souvenir2007} and computer vision \cite{Pennec2006, Pennec1997, Sen2008, Tuzel2007} produce data  in such forms. This encourages researchers to conduct the analysis directly on the manifold. %rather than in the Euclidean space or the enlarged Euclidean space using basis expansions. 
The rationale behind this is that usage of the metric on the manifold is much more reasonable than using the metric in the Euclidean space, if the data resides on manifolds.  However, the methodology defined using the manifold space for classification is still lacking. In order to perform reliable classification for data points on manifolds, a strategy for developing statistical tools, such as the non-linear classification boundary, in parallel with their Euclidean-counterparts, is significantly relevant.

Though we have seen tremendous efforts in the development of statistical procedures for classification problems, these efforts have mainly been focused on constructing the separating hyperplane between two classes in the Euclidean space. The optimality is essentially built on finding {\it linear} (affine) hyperplanes that separate the data points as well as possible. Among them, the linear discriminant analysis (LDA), or the slightly different logistic regression method, manifest themselves through the seeking of the hyperplane by minimizing the so-called discriminant function, and are thus able to trace out a linear boundary separating the different classes \cite{hastie_09esl}. Further to the linear boundary, the support vector machine (SVM) finds a seemingly different separating hyperplane; that is, the hyperplane is actually found (up to some loss function) not in its original feature space, but in an enlarged space, by transforming the feature space into an unknown space via basis functions.  Furthermore, locally linear SVM variants \cite{localSVM1, localSVM2} learn smooth classifiers from existing descriptions of manifolds that characterize the manifold as a set of piecewise affine charts and show some promise in application. There have been several pieces of works of research on the statistical methods on manifolds over the past decades, centering around finding the main mode of variation \cite{Frechet1948, Huckemann2010} in the data, or finding a manifold version of the principal components for the data \cite{Jupp1987, Fletcher2004, Huckemann2006, Kumi2007, Fletcher2007, Kenobi2010, Jung2012, Eltzner2017}, in terms of dimension reduction. Other approaches \cite{Zhu2003, Belkin2006} are more attached to regularization that exploits the geometry of the input space, achieving better generalization error. However, none of them seem to be naturally adaptable for deriving a boundary on a manifold, due to their ``non curve-fitting'' nature. A Bayesian framework \cite{Bhattacharya2012} has also been proposed to use the kernel mixture model for the joint distribution of the labels and features, with the kernel expressed in product form and dependence induced through the unknown mixing measure. The principal curve by \cite{hastie1989} is a self-consistent curve defined in Euclidean space. Its natural extension to Riemannian manifolds \cite{Hauberg2015} is based on the replacement of the Euclidean conditional mean with the intrinsic mean of the local data on manifolds. Recently, the principal flow \cite{Panaretos2014} has been proposed as a one-dimensional curve defined on the manifold, such that it attempts to follow the main direction of the data locally, while still being able to accommodate the ``curve fitting'' characteristic on the manifold. The variational principal flow \cite{level-2017} incorporates the level set method to obtain a fully implicit formulation of the problem. The principal sub-manifold \cite{prin-sub2017} extends the principal flow to a higher dimensional sub-manifold. It is natural to raise the question as follows: Is there a way for the data to be separated directly on the manifold?% without an attempt to formulate the same problem in another unknown space?%is there anything interesting to be found that can separate the data directly on the manifold?%without attempting to formulate the same problem in another unknown space?\\

%In this sense, SVM is only capable of producing the namely nonlinear boundary with respect to the original space. This being said, a direct extension of SVM while retaining its use as a classifier when it comes to the manifold is not straightforward. 
Inspired by the principal flow, we tackle the problem of finding the classification boundary endowed with a curved metric on Riemannian manifolds. We explore the limitations inherent in the problem when trying to find such a boundary. 
%We do not intend to merely search a non-linear boundary, directly on the manifold, that satisfies a certain optimisation condition with respect to the data points. 
Our idea is to trace a boundary out of the two summary curves (principal flows) from the two classes, and at the same time retain some canonical interpretation for the boundary. Our intuition is that, as the two principal flows represent the mean trend of the two classes, in order to classify the points it is enough to separate the two flows in some optimal way. This means that one does not need to consider the data points beyond the two flows on each side, as they are irrelevant to the classification if we can separate the flows well. Naturally, because of the two principal flows, the process of constructing the boundary can be supervised, in the sense that the boundary grows itself by borrowing strength from the two principal flows. To achieve this, the key insight is the margin, a measure of distance between the target boundary and the two principal flows, subject to the presence of noise originating from each class. In principle, an optimal boundary can be framed by maximizing the margin between the target boundary and the two corresponding principal flows. The optimization involved therein can be relaxed by fine-tuning the subspace of the vector field from the two principal flows, up to their parallel transport on the manifold. From this perspective, the boundary retains the characteristic of being principal, in the sense that at each point of the boundary, it points to the direction calculated over the two directions of the vector field from the two principal flows. This finally results in a classification boundary, which is named the principal boundary to draw a relation to the principal flow. %The details are to be discussed in Section \ref{sec:method}.      

\begin{figure} 
\centering
        \begin{subfigure}[b]{0.35\textwidth}
        		\centering
        		\includegraphics[width=2 in]{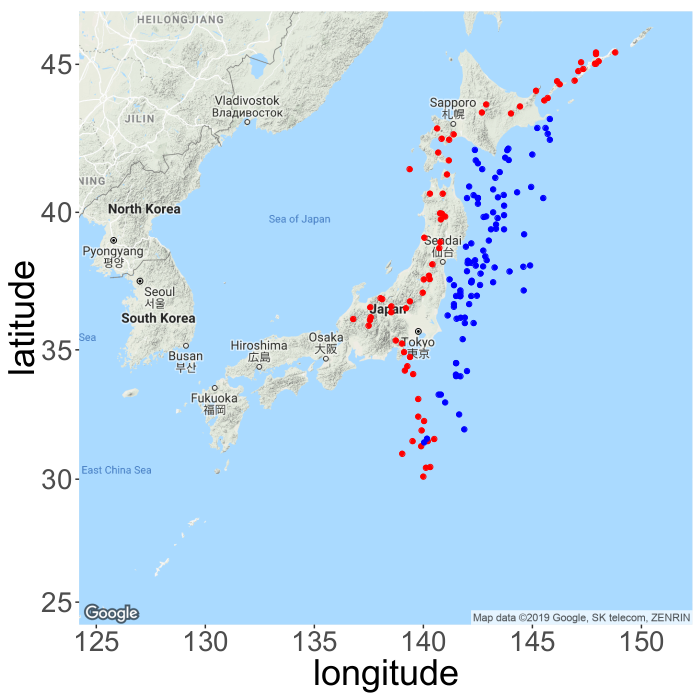}
         	\caption{}
        \end{subfigure}\\
         %\hspace{0.15 in}
	\begin{subfigure}[b]{0.35\textwidth}
                %\centering
                %\includegraphics[width=1\linewidth]{PDF_Data/flow_offshore_flat.pdf}
                %\includegraphics[width=1\linewidth]{PDF_Data/boundary_offshore_flat1.pdf}
		%\includegraphics[width=1\linewidth]{PDF_Data/boundary_offshore_flat1_new.png}
		\includegraphics[width=1\linewidth]{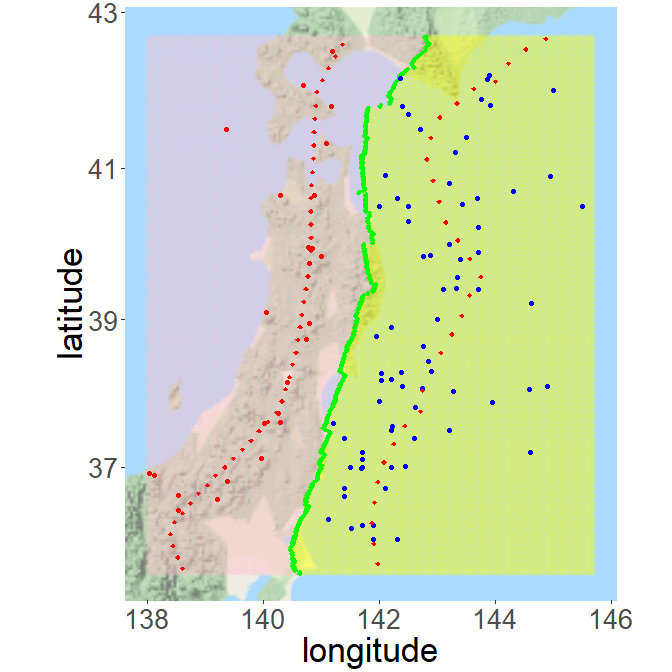}
        		\caption{}
        \end{subfigure}% 
 	\begin{subfigure}[b]{0.35\textwidth}
               % \centering
                %\includegraphics[width=1\linewidth]{PDF_Data/flow_offshore_flat.pdf}
                %\includegraphics[width=1\linewidth]{PDF_Data/boundary_offshore_flat2.pdf}
		%\includegraphics[width=1\linewidth]{PDF_Data/boundary_offshore_flat2_new.png}
		\includegraphics[width=1\linewidth]{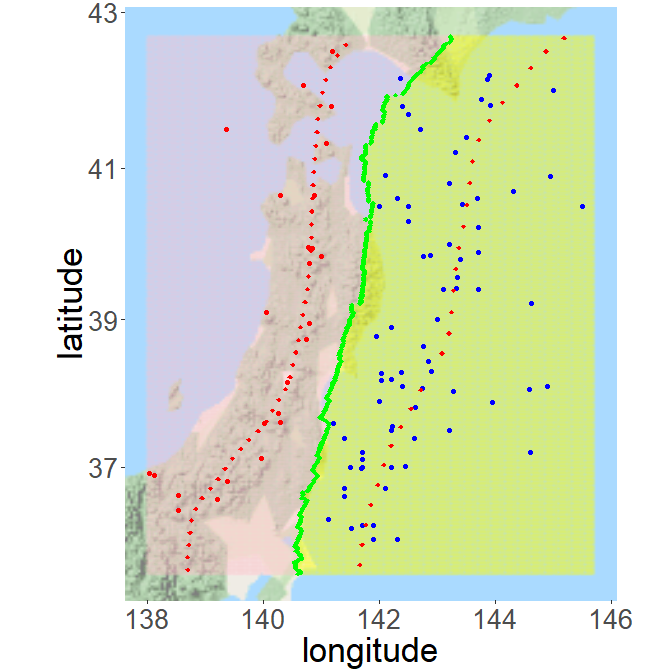}
        		\caption{}
        \end{subfigure}% 
        \begin{subfigure}[b]{0.35\textwidth}
               % \centering
                %\includegraphics[width=1\linewidth]{PDF_Data/flow_offshore_flat.pdf}
                %\includegraphics[width=1\linewidth]{PDF_Data/boundary_offshore_flat3.pdf}
		%\includegraphics[width=1\linewidth]{PDF_Data/boundary_offshore_flat3_new.png}
     \includegraphics[width=1\linewidth]{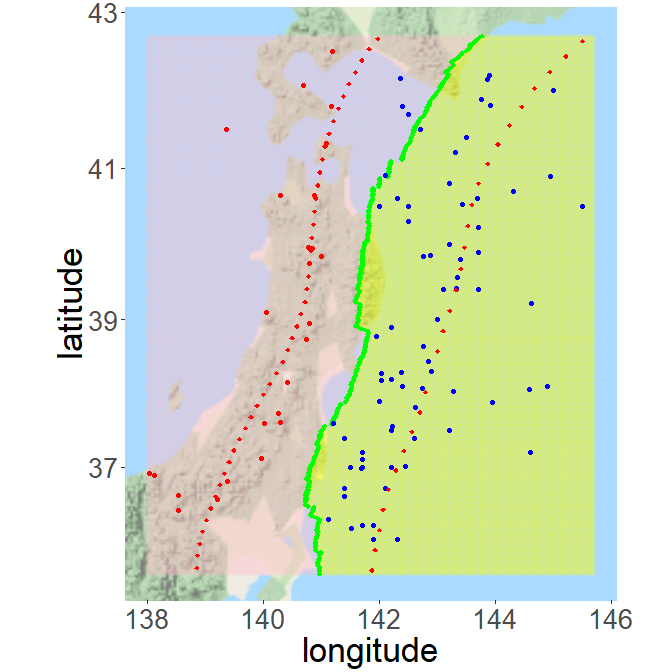}
					\caption{}
        \end{subfigure}% 
        \caption{Seismic events data of significant offshore earthquakes with magnitude 7.5 or greater (in blue) between 1900 and 2018, and major volcanoes (in red) in 2001 on a flat world atlas (a). Plots of the principal boundary (in green) between the volcano region (transparent pink) and earthquake region (transparent yellow) in the zoom-in area ($138.0\degree - 145.7\degree$ E, $35.5\degree - 42.7\degree$ N) (b)-(d). The bottom row illustrates the effect of varying flows on the classification performance of the boundary: the two clusters are correctly classified with no misclassified points (b), one misclassified earthquake point (c), and four misclassified earthquake points (d).}
\label{illustrate-boundary}
\end{figure}

Figure \ref{illustrate-boundary}  shows the data of major volcano and significant offshore earthquake activity that occurred around the region of eastern Japan. The principal boundary and misclassified points illustrate the effect of varying flows on the classification performance of the boundary. The transparent pink and transparent yellow areas (Figure \ref{illustrate-boundary}(b)-(d)) represent the classified regions that correspond to the volcano and earthquake activity respectively, contrasting with the volcano and earthquake data plotted in red and blue. The principal boundary (in green) exhibits different behavior, as the principal flows vary with respect to their locality parameters: the boundary classifies the two data clouds correctly (Figure \ref{illustrate-boundary}(b)); there is one misclassified earthquake point in the volcano region (Figure \ref{illustrate-boundary}(c)); there are three misclassified earthquake points in the volcano region, and one misclassified earthquake point on the boundary (Figure \ref{illustrate-boundary}(d)). Readers may refer to Section \ref{realdata} for the analysis. The principal boundary is able to follow the trajectory towards the direction of maximum margin between the flows, while separating the data points as best as it could.

The principal boundary should be a co-dimensional sub-manifold {\it i.e.}, a sub-manifold with dimension $m-1$ in the $m$-dimensional manifold. In this paper, we focus on 2-dimensional manifolds, {\it i.e.}, $m=2$, for simplicity, and hence, the principal boundary is also a curve in the 2-dimensional manifold. This restriction can be extended to higher dimensional manifolds with further efforts.
We demonstrate how the problem of obtaining the principal boundary can be transformed to a well-defined integration problem in Section \ref{sec:class-bound}, with a motivation and an introduction of the margin in Section \ref{margin}, and the optimal boundary  between two curves in Section \ref{opt-boundary}. The formal definition of the principal boundary is given in Section \ref{sec:class-bound}. 
%We show that on the manifold, the principal boundary reduces to the SVM boundary, at least locally; that is, the segment of the principal boundary coincides with that of SVM. 
Section \ref{svm-pd} contains the property of the principal boundary for a finite sample with a detailed analysis of the relation between the boundary and SVM. Some optimality and convergence properties of the random principal boundary, given data with random distribution on manifolds, are provided in Section \ref{random-pd}. Generally speaking, our formulation of the boundary is feasible for any Riemannian manifold provided that the geodesic is unique, although an unknown geodesic might increase the complexity of computation.%The relation has been Theorem \ref{thm:equivalence} is proved in Appendix 1. %The principal sub-manifold also provides a complementary notation to that of a principal surface by \cite{hastie1989}, as a self-consistent surface defined in Euclidean space. 

The remaining part of the paper is organized as follows. We start with a brief introduction of the principal flow (Section \ref{pflow}-\ref{modiflow}) with a modified vector field and a modified principal flow. %Section \ref{sec:method} is the Methodology section. 
Section \ref{algorithm} investigates an implementable algorithm used to determine the principal boundary. In Section \ref{sim} and \ref{realdata}, we illustrate the principal boundary by means of simulated examples and real application.  The Appendix and supplementary materials provide all the technical details. We end the paper with a discussion.

\section{The Problem}
We are interested in the following problem: let $\mathcal{M}$ be a Riemannian manifold in $\mathbb{R}^d$ with the dimension $m < d$. Let $x_{1,i}$ ($i= 1, \ldots, n_1$) be the data points on $\mathcal{M}$ with label $\ell_{1,i}=+1$ ($i= 1, \ldots, n_1$) and $x_{2,i}$ ($i= 1, \ldots, n_2$) be the data points on $\mathcal{M}$ with label $\ell_{2,i}=-1$ ($i  = 1,\ldots,n_2$). We are looking for a classification boundary, say $\gamma$, an $m-1$ sub-manifold on $\mathcal{M}$, that %is as wide as possible between the data points for class 1 and -1.
separates the data points for class 1 and -1 as well as possible. 
%In this paper, we focus on the case for $m=2$ manifolds. That is, $\gamma$ is also a curve on a two-dimensional manifold. The extension to higher manifolds will be given in a separate paper. 

The heuristic behind the principal boundary is that first we construct the two mean flows $\gamma_1$ and $\gamma_2$ of the data points $\{x_{1,i}\}$ and $\{x_{2,i}\}$, respectively. Each mean flow represents the principal direction of the data variation for each class on $\mathcal{M}$. Second, the classification problem can now be rephrased as the finding of a flow $\gamma$, lying between $\gamma_1$ and $\gamma_2$, which separates the two as well as possible.   

The two mean flows are in fact called principal flows. Before we continue, let us digress slightly and review the principal flow.

\subsection{Preliminaries}
\label{preliminary}

Let $x_i\  (i=1,\ldots,n)$ be $n$ data points on a complete Riemannian manifold $\mathcal{M}$ of dimension $m$, where $m <d$, embedded in the linear space $\mathbb{R}^d$. 

We assume that a differentiable function $F: \mathbb{R}^{d} \rightarrow \mathbb{R}^{m}$ always exists, such that
\begin{align*}
\mathcal{M} := \left\{x \in \mathbb{R}^{d}: F(x)= 0  \right\}.
\end{align*} 
 For each $x\in \mathcal{M}$ the tangent space at $x$ will be denoted by $T_x\mathcal{M}$, then $T_x\mathcal{M}$ is characterized by the equation
 \begin{align*}
 %T_x\mathcal{M} = \left\{y: DFy = 0, y \in \mathbb{R}^d \right\}.
 T_x\mathcal{M} = \left\{v\in \mathbb{R}^d: \nabla F_xv = 0  \right\}.
 \end{align*} 
 Thus, $T_{x}\mathcal{M}$ is in fact a {\it vector space}, the set of all tangent vectors to $\mathcal{M}$ at $x$, which essentially provides a local vector space approximation of the manifold $\mathcal{M}$.%which essentially provides a local vector space approximation of the manifold $\mathcal{M}$. %This is by analogue the derivative of a real-valued function that provides a local approximation of the function. Let $g$ be a smooth family of inner product associated with the manifold $\mathcal{M}$:
 %\begin{align*}
 %g_x : T_x \mathcal{M} \times T_x\mathcal{M} \rightarrow \mathbb{R}
 %\end{align*}
 %Then for any $v\in T_x\mathcal{M}$, the norm of $v$ is defined by
 %\begin{align*}
 %\|v \| = g_x(v,v).
 %\end{align*}

%\bed \label{geo}
%A curve, $\gamma: [0,\delta ] \rightarrow \mathcal{M}$ is a geodesic if and only if $\dot{\gamma}(t)$ is parallel along $\gamma$, that is, the acceleration vector
%\begin{align*}
%\frac{d (\dot{\gamma})}{dt}=0,
%\end{align*}
%where $\delta$ is sufficiently small such that the curve is well-defined. The definition means that $\frac{d (\dot{\gamma})}{dt}$ is normal to $T_{\gamma(t)}\mathcal{M}$ at any time $t$.
%\eed
By equipping the manifold with the tangent space, we define two mappings back and forth between $T_x \mathcal{M}$ and $\mathcal{M}$: 1) the exponential map, well defined in terms of geodesics, is the map:  
\begin{equation} \label{exp}
\mbox{{\bf exp}}_{x}: T_x \mathcal{M} \rightarrow \mathcal{M}%\quad \mbox{by } \mbox{{\bf exp}}_{x}(v)=\gamma_v(1)
\end{equation}
by $\mbox{{\bf exp}}_{x}(v)=  \gamma(\|v\|)$ with $\gamma$ a geodesic starting from $\gamma(0) = x$ with initial velocity $\dot{\gamma}(0) = v/\|v\|$ and $\|v\| \leq \delta$, and 2) the logarithm map (the inverse of the exponential map), is locally defined at least in
the neighborhood of $x$,  
\begin{equation} \label{log}
\mbox{{\bf log}}_x : \mathcal{M} \rightarrow T_x \mathcal{M}.
\end{equation}
Here, the {\bf exp} and {\bf log} are defined on a local neighborhood of $0$ and $x$ such that they are all well-defined, away from the cut locus of $x$ on $\mathcal{M}$.\\

Let $x, y \in \mathcal{M}$. Denote all (piecewise) smooth curves $\gamma(t): [0,1] \rightarrow \mathcal{M}$ with endpoints such that $\gamma(0)=x$ and $\gamma(1)=y$. The {\it geodesic distance} from $x$ to $y$ is defined as 
\begin{equation}\label{lencurve}
%d_{\mathcal{M}}(x,y) = \mbox{inf } s(\gamma) %d_{\mathcal{M}}(p,q)=
d_{\mathcal{M}}(x,y) = \mbox{inf } \int_{[0,1]} \left\|\dot{\gamma}(t)\right\|dt.
\end{equation}
%where $s(\gamma)=\int_{[0,1]} \left\|\dot{\gamma}(t)\right\|dt$. %\int_{[0,1]} \langle \dot{\gamma}(t), \dot{\gamma}(t) \rangle ^{\frac{1}{2}} dt$
% is computed by integrating the $\dot{\gamma}(t)$ along the curve,
Minimizing \eqref{lencurve} yields the shortest distance within $\mathcal{M}$ between the two points $x$ and $y$.

\subsection{Definition of principal flow}
Technically, the principal flow incorporates two ingredients: a local covariance matrix and a vector field. 
For the data point $x_j$, choose a neighborhood ${\cal N}(x_j,h)$ of $x_j$ with a radius $h$, defined as
\[
	{\cal N}(x_j,h) = \{x_i: d_{\cal M}(x_i,x_j)\leq h\}.
\] 
Accordingly, the local covariance matrix is defined as
$$
\Sigma_h(x_j)=\frac{1}{\sum_{i}\kappa_h(x_i, x_j)} { \sum_{i}}\mbox{{\bf log}}_{x_j}(x_i)  \otimes \mbox{{\bf \mbox{log}}}_{x_j}(x_i) \kappa_h(x_i,x_j)
$$
where $ y\otimes y:= y y^T$, $\kappa_h(x_i,x_j)=K(h^{-1} d_\mathcal{M}(x_i, x_j))$ with a smooth non-increasing uni-variate kernel $K$ on $[0,\infty)$. %We remark here that all the above definitions are under the assumption that the first and second eigenvalues of $\Sigma_h (x_j)$ are distinct.

Let $B \subset \mathcal{M}$ be a connected open set covering $x_{i}$ ($i= 1, \ldots, n$) and such that $\mbox{{\bf log}}_x y$ is well defined for all $x, y \in B$. Assume that $\Sigma_h(x_j)$ has distinct first and second eigenvalues for all $x \in B$. The vector field is defined in the way that the first eigenvector $e_1(x_j)$ (or eigenvalue $\lambda(x_j)$) of $\Sigma_h(x_j)$ is extended to a vector field $W:= \{W(x)\equiv v(x_j): x \in \mathcal{N}(x_j, h)\}$ where $v(x_j)=e_1(x_j)$; that is, for any $x \in  \mathcal{N}(x_j, h)$, we have
\begin{eqnarray} \label{vectorfield} 
\Sigma_{h}(x) W(x)= \lambda(x) W(x) (\mbox{i.e., } W(x) \in W).
\end{eqnarray} 
In the meantime, it has been proved \cite{Panaretos2014} that $W: \mathcal{N}(x_j,h) \rightarrow \mathbb{R}^d$ is a differentiable mapping with $W(x)$ independent of the local coordinates of the tangent space $T_x\mathcal{M}$.

\label{pflow}
\bed \label{flow}
The principal flow $\gamma$ of $x_{i}$ ($i= 1, \ldots, n$) is defined as the union of the two curves, $\gamma^+$ and $\gamma^-$, satisfying the following two variational problems respectively 
\begin{eqnarray} 
%\gamma^+ &=& \arg \sup_{\gamma \in \Gamma (x_0,v_0)} \int_{0}^{\ell(\gamma)} \left\langle \dot{\gamma}(t), W(\gamma(t))\right\rangle dt\\
%\gamma^- &=& \arg \inf_{\gamma \in \Gamma (x_0,-v_0)} \int_{0}^{\ell(\gamma)} \left\langle \dot{\gamma}(t), W(\gamma(t))\right\rangle dt
\gamma^+ &=& \arg \sup_{\gamma \in \Gamma (x_0,v_0)} \int_{\gamma} \left\langle \dot{\gamma}, W(\gamma)\right\rangle ds\\
\gamma^- &=& \arg \inf_{\gamma \in \Gamma (x_0,-v_0)} \int_{\gamma} \left\langle \dot{\gamma}, W(\gamma)\right\rangle ds
\end{eqnarray}
\eed
where $x_0$ is the starting point and $v_0$ is a unit vector at $x_0$. The point $x_0$ can be chosen as the Fr\'{e}chet mean $\bar{x}$ of $x_{i}$ ($i= 1, \ldots, n$) such that
\begin{align*}
\bar{x}= \margmin_{x \in \mathcal{M}} \frac{1}{n} \sum_{i=1}^{n} d^2_{\mathcal{M}}(x, x_i),
\end{align*}
%where $d_{\cal M}(\cdot, \cdot)$ is the geodesic distance between two points on $\cal M$, 
or any other point of interest. 
%The above integral is defined over the class 
%\begin{multline} \label{candidate-flow}
% \Gamma(x_0, v_0)= \Big\{\gamma:[0,r] \rightarrow \mathcal{M}, \gamma \in C^2(\mathcal{M}), \gamma(t) \neq \gamma(t') \mbox{ for } t\neq t',\\
%                \gamma(0)=x_0, \dot{\gamma}(0)=v_0, \ell(\gamma[0,t])=t \mbox{  for all  } 0 \leq t \leq r \leq 1 \Big\}.
%\end{multline}
%where $\ell(\gamma)$ is the length of $\gamma$.  
Note that $\Gamma(x_0, v_0)$ is the set of all non-intersecting differentiable curves on $\mathcal{M}$.

It can be seen that the curve $\gamma^+$ starts at $x_0$ and follows the direction of the vector field, and the curve $\gamma^-$ starts at $x_0$ and goes in the opposite direction of the vector field. Thus, the integral for $\gamma^-$ is negative, which explains why the infimum appears in its definition.

Under the principal flow definition, we can define $\gamma_1$ of $x_{1,i}$ ($i= 1, \ldots, n_1$) as the union of $\gamma_1^+$ and $\gamma_1^-$, and $\gamma_2$ of $x_{2,i}$ ($i= 1, \ldots, n_2$) as the union of $\gamma_2^+$ and $\gamma_2^-$.  For convenience, we will only consider the flow $\gamma_1^+$ in (1) of Definition \eqref{flow}, and re-name it as $\gamma_1$. By symmetry, the solution to the flow $\gamma_1^-$ in (2) of Definition \eqref{flow} can be carried out in the same way. Similarly, we will restrict the discussion to $\gamma_2^+$ and re-name it as $\gamma_2$.

\subsection{Modified principal flow}
\label{modiflow}
The principal flow relies heavily on the vector field. However, the original definition (see \eqref{vectorfield}) of the vector field strictly constructs the direction of every point in the field to  solely point to the eigenvector of the local covariance matrix. This definition could be problematic when we need a more delicate field for the flow. To be exact, for each point belonging to a neighborhood where the field is calculated, it can also be in other neighborhoods. It turns out that we will need to modify the vector field for the principal boundary, since the vector field plays a rather crucial role in the problem.

We will equip a vector field for each training sample (say $x_j$) in the data $\{ x_i \} (i=1, \cdots, n)$. For samples in each neighborhood, say ${\cal N}(x_i,h)$, determine a locally dominate or principal vector $v_i$ through the local tangent PCA. 

A sample $x_j$ can be the neighbor of multiple points. Let $I_j$ be the index set of neighbor sets ${\cal N}(x_i,h)$ that holds $x_j$ as a neighbor. The modification of the vector field amounts to the overall effect of holding multiple neighborhoods for a point. To achieve this, it is very natural to equip a vector for $x_j$ as a weighted sum of the locally principal vectors $\{v_i: i\in I_j\}$. Let $c_i$ be the mean of ${\cal N}(x_i,h)$ when we determine $v_i$. Then, we assign the vector $v(x_j)$ for $x_j$, which is the projection $v_{\cal M}(x_j)$ of the weighted sum
\[
	v(x_j) = \sum_{i\in I_j}w_{ij}v_i, \quad w_{ij} = \frac{\exp(-d_{\cal M}(x_j,c_i))}{\sum_{i\in I_j} \exp(-d_{\cal M}(x_j,c_i))}
\]
onto the manifold at $x_j$. 

As soon as we have a vector field constructed above, a principal flow $\gamma$ of the given data set $\{ x_i\}$ is defined by
\begin{eqnarray}\label{mflow} 
	\gamma 
	= \arg \sup_{\gamma \in \Gamma (x_0,v_0)} \int_{\gamma} \left\langle \dot{\gamma}, \sum_{x_j\in {\cal N}(p, h)}v(x_j)\right\rangle ds
\end{eqnarray}
That is,  at the point $p\in\gamma$, the tangent $\dot\gamma$ should match the vector field of samples in the neighborhood ${\cal N}(p, h)$. To differentiate from the principal flow in Definition \ref{flow}, we call $\gamma$ the modified principal flow.

\section{Methodology}
\label{sec:method}
Now we return to the original question, that is: Given two principal flows $\gamma_1$ and $\gamma_2$ determined from two data sets $\{x_{1,i}\}$ and $\{x_{2,j}\}$, can we find a %\footnote{In principal, $\gamma$ can be of any dimension $m-1$; here, we restricted it in the setting where $m=2$.} 
$\gamma$, that can be used as a classification boundary between the two classes?  In principle, $\gamma$ should be co-dimension one (of dimension $m-1$); for simplicity, here we assume that $m=2$, and hence, $\gamma$ is also a flow in the 2-dimensional manifold.

%\begin{multline} \label{candidate}
% \Gamma(x_c, v_c)= \Big\{\gamma:[0,r] \rightarrow \mathcal{M}, \gamma \in C^2(\mathcal{M}), \gamma(t) \neq \gamma(t') \mbox{ for } t\neq t',\\
%                \gamma(0)=x_c, \dot{\gamma}(0)=v_c, \ell(\gamma[0,t])=t \mbox{  for all  } 0 \leq t \leq r \leq 1 \Big\}.
%\end{multline}
%where $\gamma(0)=x_c$ and $\dot{\gamma}(0)=v_c$ are initial conditions for $\gamma$ and $\ell(\gamma)$ is the length of $\gamma$. 
%Therefore, by saying the classification boundary, we refer to the best  one. Although the definition is given in the next Section, we motivate the general idea of constructing such a boundary here.   
 
Strictly speaking, many $\gamma$s in $\Gamma(x_0, v_0)$  that separate two classes of data could exist. Therefore, by using the term the classification boundary, we refer to the best one. We present the general idea of constructing such a boundary here, leaving the formal introduction in Section \ref{sec:class-bound}: let the $\gamma$ start from $\gamma(0)$ and move infinitesimally in the direction of $\dot{\gamma}(0)$. At this moment, we assume that both $\gamma(0)$ and $\dot{\gamma}(0)$ are carefully chosen so that the flow moves in the correct direction. Once the first move has been made, it may no longer make sense to continue moving in the same direction $\dot{\gamma}(0)$. One may ask, then, in which direction should we move towards? Obviously, we should not move towards $\gamma_1$ or $\gamma_2$, since this would cause $\gamma$ to move close to either of the two flows. To update the direction, a natural strategy that plays an important role in building the boundary is to let  $\gamma$ move in a direction supervised by $\gamma_1$ and $\gamma_2$; that is, we follow the vector field inherited from $\gamma_1$ and $\gamma_2$, then move by choosing the proportional amount of vector field from $\gamma_1$ and $\gamma_2$  each time. Indeed, the right amount of vector field to choose for the next move is essentially an optimization problem, the derivation of which will be discussed in the next section. This being said, the intuitive version of such a boundary is not unique in the sense that a parallel curve satisfying the same condition always exists, and this can be seen by varying the initial point. To achieve the classification, let us view the problem slightly differently: note that only the points lying in between $\gamma_1$ and $\gamma_2$ could influence $\gamma$, so a very straightforward approach is to choose the tangent vector for the next move along the direction that creates the biggest margin between the data points for class $+1$ and $-1$. Under this rationale, iterating the process would approximately  trace out an integral curve that is not only proportionally compatible to the vector fields of the two flows at each point, but also, more importantly,  separates the margin, therefore allowing it to be considered as a classification boundary. 

%We will now make sense the original problem as an optimization problem. The rigorous definition will involve a modification of the vector field and a definition of the margin.

\subsection{Margin}
\label{margin}
At each point $p\in \gamma$, the tangent vector of $\gamma$ at $p$ should be the locally principal vector of samples in ${\cal N}(p, h)$. Suppose the distinct first and second largest eigenvalues of the local covariance matrix of the centered samples for ${\cal N}(p, h)$ exist. The principal vector is the dominant eigenvector in ${\cal N}(p, h)$, corresponding to its largest eigenvalue $\lambda_1$.  
The local PCA also determines the second largest eigenvalue value $\lambda_2$. The ratio
\[
	\sigma_\gamma(p) = \frac{\lambda_2}{\lambda_1}h
\] 
approximately indicates the largest distance of samples in the neighborhood ${\cal N}(p, h)$ deviated from the mean along the eigenvector corresponding to $\lambda_2$.

The distance of a point $q$ of the manifold to the principal flow $\gamma$ is defined as
\begin{equation}\label{projection}
\mathpzc{d}_{\cal M}(q,\gamma) = \inf_{p\in \gamma} d_{\cal M}(q, p),
\end{equation}
where $d_{\cal M}(p, q)$ is the geodesic distance between $p$ and $q$ on $\cal M$. We assume that the distance $\mathpzc{d}_{\cal M}(q,\gamma)$ is achievable, i.e., there is a point $p\in \gamma$ such that $d_{\cal M}(q,p) = \mathpzc{d}_{\cal M}(q, \gamma)$. 
%Furthermore, we assume that the minimiser is unique. We call $p$ as the projection of $q$ onto $\gamma$, and denote it as $p= p_{\gamma}(q)$ as a function of $q$.  
Such a minimizer $p$ is referred as a projection of $q$ onto $\gamma$. We denote by ${\cal P}_{\gamma}(q)$ as the set of all the minimizers. 
%For simplicity, we assume that the minimiser is unique. Otherwise, the function $\sigma_\gamma$ defined below should be the supremum over the set of minimisers.

Hence, if $\mathpzc{d}_{\cal M}(q,\gamma)>\sigma_{q,\gamma} = \sup_{p\in{\cal P}_{\gamma}(q)} \sigma_\gamma(p)$,% with $p=p_\gamma(q)$, 
the gap
\[
	m_\gamma(q) = \mathpzc{d}_{\cal M}(q,\gamma) - \sigma_{q,\gamma}
\]
is a ``soft'' margin of $q$ for classifying $\gamma$ in the sense that the neighbor set ${\cal N}(p,h)$ locates in a side of $q$ for each $p\in{\cal P}_{\gamma}(q)$, at least locally. More generally, given curve $\gamma'$ on the same manifold, if we always have a positive value of $m_\gamma(q)$ for each $q\in \gamma'$, then $\gamma'$ is located on one side. For simplicity, in the following discussion, we always assume that the projection $p$ of $q$ onto $\gamma$ is unique and denote it as $p= p_{\gamma}(q)$ as a function of $q$.

\subsection{Optimal boundary}
\label{opt-boundary}

Before describing the behavior of the best $\gamma$ in a finite sample setting, let us first consider the phenomenon of an optimal boundary. 

\begin{condition}{(Optimal boundary conditions)} \label{Prop:optimal-boundary} consider any curve $\gamma$ on $\mathcal{M}$ that could be potentially used as a boundary, located between $\gamma_1$ and $\gamma_2$.
\begin{itemize}
\item[1.]  For any %point 
$q\in \gamma$, the projection of $q$ to the two flows, $\mathpzc{d}_{\cal M}(q,\gamma_1)$ and $\mathpzc{d}_{\cal M}(q,\gamma_2)$, is achievable.  %The segment is discretized as $\gamma_c^*= \{q_1^*, \cdots,q_{M}^*\}$, where $q_j^*  \in \mathcal{M}$.
%\item[2.] For the two projected points $p_1=p_{\gamma_1}(q)$ and $p_2=p_{\gamma_2}(q)$, we always have zero gaps; that is, $\sigma_{\gamma_1}(p_1)=0$ and $\sigma_{\gamma_2}(p_2)=0$. 
\item [2.] All $\gamma$s are smooth and have equal length.
\end{itemize}
\end{condition}

%\bed \label{optimal-boundary}
%A unit-speed curve $\widetilde{\gamma} \in \Gamma$ is called the optimal boundary if it maximises the integral of additive distance $\mathpzc{d}_{\cal M}(q,\gamma_1) + \mathpzc{d}_{\cal M}(q,\gamma_2)$ over $\gamma$. That is, 
%\begin{align}\label{opt-2eq}
%	%\gamma_c = \arg\max_{\gamma\in\Gamma} \int_\gamma \min \big\{m_{\gamma_1}(\gamma(t)), m_{\gamma_2}(\gamma(t))\big\}dt.
%	\widetilde{\gamma} =& \arg\max_{\gamma\in\Gamma} \int_\gamma  \left(\mathpzc{d}_{\cal M}(q,\gamma_1) + \mathpzc{d}_{\cal M}(q,\gamma_2)\right)ds.\nonumber \\
%	         =& \arg\max_{\gamma\in\Gamma} \int_\gamma  \min \big\{\mathpzc{d}_{\cal M}(q,\gamma_1), \mathpzc{d}_{\cal M}(q,\gamma_2)\big\}ds.
%\end{align}
%\eed
\bed \label{optimal-boundary}
A unit-speed curve $\widetilde{\gamma} \in \Gamma$ is called the optimal boundary if it maximizes the integral of additive distance $\mathpzc{d}_{\cal M}(q,\gamma_1) + \mathpzc{d}_{\cal M}(q,\gamma_2)$ over $\gamma$. That is, 
\begin{align}\label{opt-2eq}
	%\gamma_c = \arg\max_{\gamma\in\Gamma} \int_\gamma \min \big\{m_{\gamma_1}(\gamma(t)), m_{\gamma_2}(\gamma(t))\big\}dt.
	\widetilde{\gamma} =& \arg\max_{\gamma\in\Gamma} \int_\gamma  \left(\mathpzc{d}_{\cal M}(q,\gamma_1) + \mathpzc{d}_{\cal M}(q,\gamma_2)\right)ds.\nonumber \\
	         =& \arg\max_{\gamma\in\Gamma} \int_\gamma  \min \big\{\mathpzc{d}_{\cal M}(q,\gamma_1), \mathpzc{d}_{\cal M}(q,\gamma_2)\big\}ds.
\end{align}
\eed

{\bf Remark}: Our treatment is abstract, in the sense that $\gamma_1$ and $\gamma_2$ can best represent the mean field of the two classes. In our view, the optimal boundary is concerned only with  $\gamma_1$ and $\gamma_2$. Note that the maximum of \eqref{opt-2eq} is attained when for any $q \in \gamma$, we have $\mathpzc{d}_{\cal M}(q,\gamma_1)=\mathpzc{d}_{\cal M}(q,\gamma_2)$.

\begin{lemma} \label{lem:opt-boundary-svm}
Assume that $\mathcal{M}=\mathbb{R}^d$ and let $\widetilde{\gamma}$ be the optimal boundary for $\gamma_1$ and $\gamma_2$. Denote that for any $q \in \widetilde{\gamma}$, $p_1=p_{\gamma_1}(q)$ and $p_2=p_{\gamma_2}(q)$ are the two projections. Let $T_{q}\widetilde{\gamma}$ be the gradient line of %$q$ on $\widetilde{\gamma}$, 
of $\widetilde{\gamma}$ at $q$,
and let $L_{\mbox{svm}}$ be the straight line determined by the SVM for $p_1$ and $p_2$. If Condition \ref{optimal-boundary} is satisfied, then $\widetilde{\gamma}$ and $L_{\mbox{svm}}$ are locally equivalent; that is, $T_{q}\widetilde{\gamma}$ coincides with $L_{\mbox{svm}}$, where $L_{\mbox{svm}}$ is through the point $q$.
\end{lemma}

 We remark that, to our knowledge, no SVM has been directly defined on manifolds. It is worthwhile to note that in the Euclidean space the optimal boundary locally reduces to the usual SVM. Lemma \ref{lem:opt-boundary-svm} is proved in Appendix A \cite{Boundarysupp}.

\subsection{Principal boundary}
\label{sec:class-bound}
Suppose that $\gamma_1$ and $\gamma_2$ are determined from the two data sets $\{x_{1,i}\}$ and $\{x_{2,j}\}$ respectively. We say $\gamma_1$ and $\gamma_2$ are separated if there is a curve $\gamma$ such that $\gamma_1$ and $\gamma_2$ are on different sides of $\gamma$, conditioning on the margins to the two curves; that is, $m_{\gamma_1}(q)>0$ and $m_{\gamma_2}(q)>0$ for all $q\in \gamma$. Clearly, such a curve $\gamma$ can correctly classify the two data sets. The minimum of $m_{\gamma_1}(q)$ and $m_{\gamma_2}(q)$ is referred as the margin of $\gamma$ to $\gamma_1$ and $\gamma_2$ at $q$, and let 
%For a point $q\in {\cal M}$ locating between $\gamma_1$ and $\gamma_2$, we call the minimum of $m_{\gamma_1}(q)$ and $m_{\gamma_2}(q)$ is the margin of $q$ with respect to $\gamma_1$ and $\gamma_2$, i.e., 
\[
	m_{\gamma_1,\gamma_2}(q) = \min\big\{m_{\gamma_1}(q),\ m_{\gamma_2}(q)\big\}.
\]

Let $\Gamma$ be the set of the classification curve with unit speed. A good classification curve should have the margin $m_{\gamma_1,\gamma_2}(q)$ to the two principal flows, as large as possible at each $p \in \gamma$. We call the ideal classification curve that has the largest value of $m_{\gamma_1,\gamma_2}(q)$ the  {\it principal boundary} of the two data sets. 

\bed \label{classification-boundary}
A unit-speed curve $\gamma_c\in \Gamma$ is called  the principal boundary if it maximizes the integral of the margin $m_{\gamma_1,\gamma_2}(q)$ over $\gamma$. That is, 
\[
	%\gamma_c = \arg\max_{\gamma\in\Gamma} \int_\gamma \min \big\{m_{\gamma_1}(\gamma(t)), m_{\gamma_2}(\gamma(t))\big\}dt.
	\gamma_c = \arg\max_{\gamma\in\Gamma} \int_\gamma \min \big\{m_{\gamma_1}(\gamma), m_{\gamma_2}(\gamma)\big\}ds.
\]
\eed

{\bf Remark}: Naturally, given the margins, we may define two smooth margin flows, $\gamma'_1$ and $\gamma'_2$, such that for any $q_1 \in \gamma'_1$, there is a $q \in \gamma_c$ satisfying %we have
$
{d}_{\cal M}(q, q_1)=m_{\gamma_1}(q). 
%\mbox{ and }\mathpzc{d}_{\cal M}(q, q_2)=m_{\gamma_2}(q),
$
Likewise, the same applies for $\gamma'_2$.%so is that for $\gamma'_2$.}
%where $q \in \gamma_c$. 
Taking the viewpoint of Definition \ref{optimal-boundary}, the principal boundary $\gamma_c$ is equivalent to the optimal boundary of $\gamma'_1$ and $\gamma'_2$. The proof is obvious, so we omit it.

Definition \ref{classification-boundary} seems to suggest optimizing the ideal principal boundary that separates two principal flows by maximizing the margin over a class of flows $\Gamma$. Theoretically, the class set $\Gamma$ contains all the curves $\gamma$ that lie in between $\gamma_1$ and $\gamma_2$. This is a much broader class than necessary to sort out a boundary from a classification point of view. Our greater interest lies in achieving such a $\gamma_c$ over a smaller set, which is more likely to be accessible. Ordinarily, to find such a set one may consider the alignment between the target principal boundary and the principal flows. Hence, we will restrict our attention to a class set $\Gamma'$, where the correspondence among $\gamma$ and $\gamma_1$ or $\gamma_2$ can be explained by a corresponding geodesic. 
%The main idea of constructing such a boundary is that, if $\gamma_c$ separates the two summarized principal flows $\gamma_1$ and $\gamma_2$ well, it separates the data points well. The key is to utilize %the margin between $\gamma_1$ and $\gamma_2$ and to explore the weights $\lambda$ in combining the vector field.

Assume the projections $p_1 = p_{\gamma_1}(q)$ and $p_2 = p_{\gamma_2}(q)$ are one-to-one for $q\in\gamma_c$; that is, a different $q$ yields a different projection onto $\gamma_1$ or $\gamma_2$. Hence, the geodesic curve ${\cal C}(p_1,p_2)$ between its two projections $p_1$ and $p_2$ must go across the principal boundary $\gamma_c$ at the original $q$. The maximization in Definition \ref{classification-boundary} means that $q\in\gamma_c$ should also be the middle point of the geodesic curve,  i.e.,
\[
	q = \arg\max_{q'\in {\cal C}(p_1,p_2)}m_{\gamma_1,\gamma_2}(q'),
\]
if both $p_1$ and $p_2$ have been pre-determined. Hence, we can equivalently define the principal boundary in a more direct way, as follows.

\bed \label{def-class-boundary}
A curve $\gamma_c$ on $\cal M$ is called the principal boundary of two principal flows $\gamma_1$ and $\gamma_2$, if any $q\in \gamma_c$ satisfies 
\begin{itemize}
	\item[(1)] the geodesic curve ${\cal C}(p_1,p_2)$ between its two projections $p_1=p_{\gamma_1}(q)$ and $p_2=p_{\gamma_2}(q)$ onto $\gamma_1$ and $\gamma_2$ also contains the point $q$, and 
	\item[(2)] $ m_{\gamma_1,\gamma_2}(q) = \max_{q'\in {\cal C}(p_1,p_2)}m_{\gamma_1,\gamma_2}(q')$. 
\end{itemize}
\eed

\subsection{Parameterization}

The condition (2) is equivalent to $m_{\gamma_1}(q) = m_{\gamma_2}(q)$ for any $q\in {\cal C}(p_1,p_2)$. 
Now let us discuss how to obtain such a $\gamma_c =\{q(t) : t\in [0,T]\}$
as a parameterized flow
\[
	\dot q(t) = v(t),\quad  t\in [0,T],
\]
starting at an initial point $q(0)$ that satisfies the condition (2) in Definition \ref{def-class-boundary}. The tangent vector $v(t)$ will be carefully chosen, as follows. 

Since the projections $p_i(t)= p_{\gamma_i}(q(t))$ can be parameterized as 
\[
	p_1(t) = p_{\gamma_1}(q(t)),\quad
	p_2(t) = p_{\gamma_2}(q(t))
\]
the principal flows can also be parameterised as  
\[
	\gamma_1 = \{ p_1(t),\  t\in [0,T]\},\quad
	\gamma_2 = \{ p_2(t),\  t\in [0,T]\}.
\] 
Hence, we are equipped with two tangent vectors $v_1(t) = \dot p_1(t)$ and $v_2(t) = \dot p_2(t)$ at $p_1(t)$ and $p_2(t)$, respectively. Numerically, the tangent vector $v_1(t)$ or $v_2(t)$ can be estimated by the vector field $W(p_1(t))$ of $\gamma_1$ at $p_1(t)$ or the vector field $W(p_2(t))$ of $\gamma_2$ at $p_2(t)$. %$W(\gamma_1(t))$ of $\gamma_1$ at $\gamma_1(t)$ or the vector field $W(\gamma_2(t))$ of $\gamma_2$ at $\gamma_2(t)$. 

The two tangent vectors $v_1(t)$ and $v_2(t)$ (or their estimates) may not necessarily lie on the same tangent plane at $q(t)$ of $\gamma_c$. A natural solution is to move the tangent vectors towards the tangent plane at $q(t)$ under a parallel transport along the geodesic curves. Let $\tilde v_1(t)$ and $\tilde v_2(t)$ be the transported tangent vectors of $v_1(t)$ and $v_2(t)$ respectively. In Appendix B \cite{Boundarysupp}, we provide the details of the machinery {\it Schild's Ladder} for an approximate implementation of the parallel transport.

As soon as the parallel transport is done at the current $q(t)$, the choice of $v(t)$ is two-fold: 1) if $v(t)$ lies in the plane spanned by $\tilde v_1(t)$ and $\tilde v_2(t)$, then there is a $\lambda(t)$ satisfying the equation 
$ 
	v(t) = \lambda(t)\tilde v_1(t)+  (1-\lambda(t)) \tilde v_2(t);
$
2) otherwise, $v(t)\approx \lambda(t)\tilde v_1(t)+  (1-\lambda(t)) \tilde v_2(t)$ with 
\[
	\lambda(t) = \arg\min_\lambda \|v(t) - (\lambda \tilde v_1(t)+  (1-\lambda ) \tilde v_2(t))\|_2,
\]
where $\| \cdot \|_2$ is the $L_2$ norm.

Although the above discussion does not immediately yield an implementable estimation of the true vector $v(t)$ ($\lambda(t)$ is not available), it gives an updating rule of estimating $v(t)$, as follows. Prior to $v(t)$,  we choose $v(t-\delta)$ with a small $\delta>0$ and estimate $\lambda(t)$ as 
\begin{align} \label{lambda_delta_t}
\lambda_\delta(t) = \arg\min_\lambda \|v(t-\delta) - (\lambda \tilde v_1(t)+  (1-\lambda ) \tilde v_2(t))\|_2.
\end{align}
Then, we check if the estimate $v_\delta(t) = \lambda_\delta(t)\tilde v_1(t)+  (1-\lambda_\delta(t)) \tilde v_2(t)$ is acceptable by testing whether $q_\delta(t)$, the projection of $q(t-\delta)+\delta v_\delta(t)$ onto the manifold, satisfies the conditions in Definition \ref{def-class-boundary} under a given accuracy. If this is not the case, we slightly tune $\lambda_\delta(t)$ to $\lambda_\delta(t):=\lambda_\delta(t)\pm\epsilon$, and check again until convergence. The updating is 
\begin{align}\label{update}
	\lambda_\delta(t):= \left\{\begin{array}{ll}
		\lambda_\delta(t)+\epsilon& {\rm if}\  m_{\gamma_1}(q_\delta(t))<m_{\gamma_2}(q_\delta(t));\\
		\lambda_\delta(t)-\epsilon& {\rm otherwise}.\\
	\end{array}\right.
\end{align}
Initially, when the tangent vector $v(0-\delta)$ is not available for determining the initial $\lambda_\delta(0)$, we can simply choose $\lambda_\delta(0) = 1/2$. 
In the next section, we will present a detailed algorithm for computing the principal boundary.

\section{Algorithm}
\label{algorithm}
In practice, finding the principal boundary can be more challenging than finding a principal flow, in the sense that the former problem is more attached to the picking of the points on the boundary. Recalling that for a point $q\in \gamma_c$, 
\[
	 \min \big\{m_{\gamma_1}(q), m_{\gamma_2}(q)\big\}
	 =\frac{1}{2}\big\{|m_{\gamma_1}(q)- m_{\gamma_2}(q)|+m_{\gamma_1}(q) + m_{\gamma_2}(q)\big\}
\]
the minimum is achieved if and only if $|m_{\gamma_1}(q)- m_{\gamma_2}(q)|$ is as small as possible. However, one cannot simply identify a sequence of such $q$'s between the two flows to form an approximate boundary. The main reason is that we require the boundary to be a smooth curve. In this respect, we need a much more sophisticated mechanism to guarantee an equal margin on both sides of the boundary, including the choice of the initial point, whereas in the case of principal flow, the picking of the initial point can be very flexible. This is particularly true when the margins differ significantly between the two classes. In these cases, picking the mean or any symmetry of the data points is no longer meaningful. To facilitate the process of generating the boundary, we will require an initial point and then a process of fine-tuning the vector field along the way, which essentially has a direct impact on the principal boundary between the two flows.

\begin{figure} 
          \includegraphics[width=0.5 \linewidth]{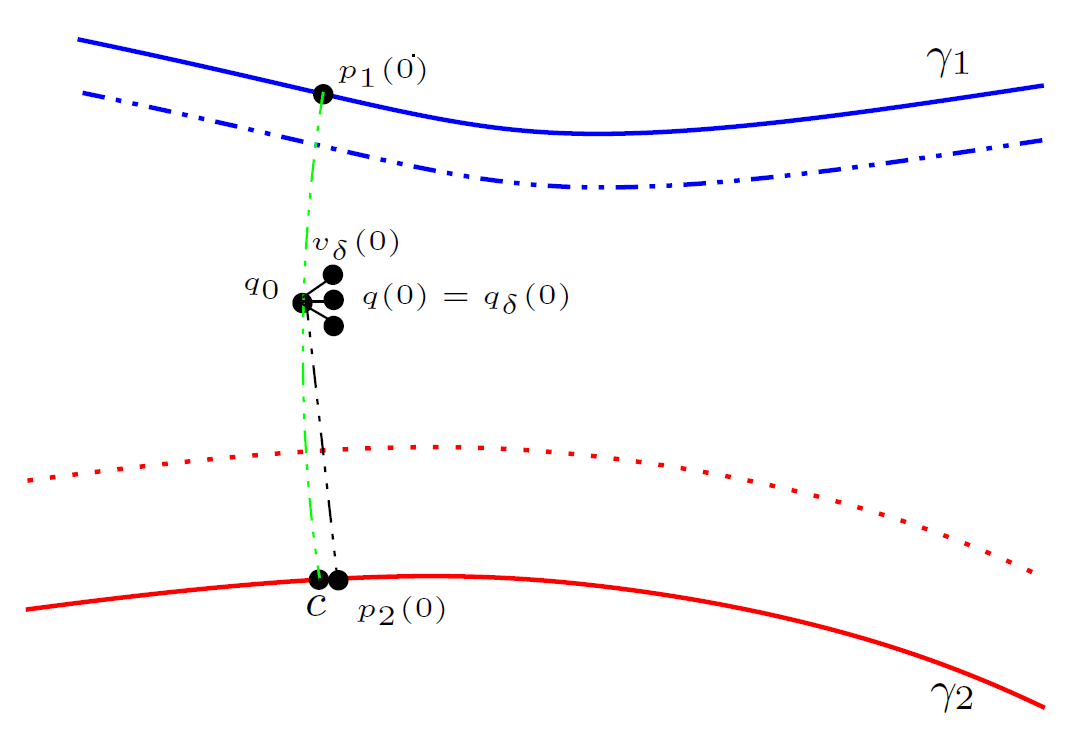}
					%\quad \quad
					\includegraphics[width=.5 \linewidth]{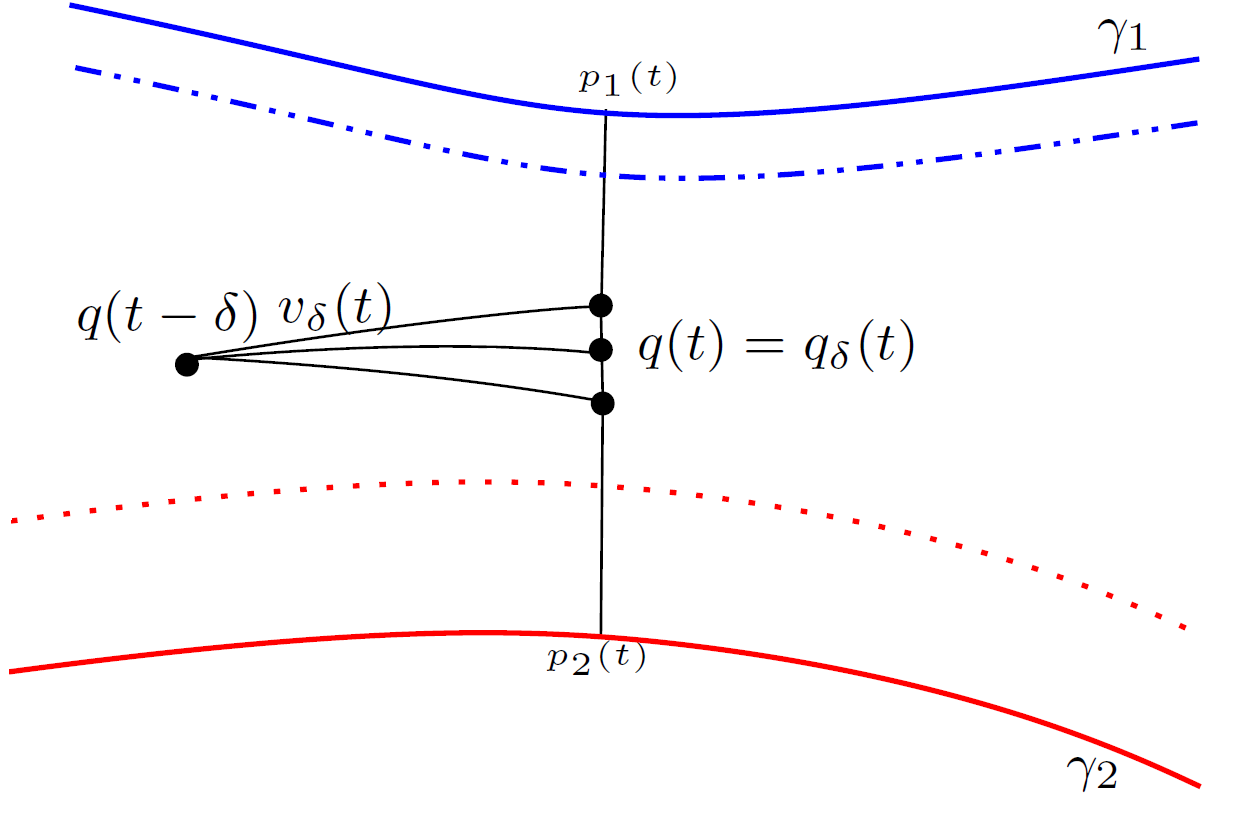}
			     \caption{Algorithm. (a) Step 1 (Initializing the boundary): $q_0$ is the warm start for finding the matching pair $p_1(0)$ and $p_2(0)$; $q(0)$ is the initial point chosen from the projection $q_{\delta}(0)=\mbox{ {\bf exp}}_{q_0}(q_0+\delta v_{\delta}(0))$ satisfying the conditions of Definition \ref{def-class-boundary}, via alternating $v_{\delta}(0)$. (b) Step 2 (Updating the boundary): $q(t-\delta)$ is used to find $q(t)$; $q(t)$ is chosen from the projection $q_{\delta}(t) = \mbox{ {\bf exp}}_{q(t-\delta)}(q(t-\delta)+\delta v_{\delta}(t))$ satisfying the conditions of Definition \ref{def-class-boundary},  via alternating $v_{\delta}(t)$.}
					\label{corealgo}
\end{figure}

We will now present a high-level description of the algorithm for computing the boundary (see Figure \ref{corealgo}), the core of which is elaborated as follows.

Step 1 ({\it Initializing the boundary}): The initialization involves finding a matching pair on $\gamma_1$ and $\gamma_2$, and calculating an initial point $q(0)$.
%(i)  Pairing: 
Arbitrarily choose a point $c\in \gamma_2$, and let $p_0' = p_{\gamma_1}(c)$ be the projection of $c$ onto $\gamma_1$. Consider the geodesic curve $g(c,p_0')$ between $c$ and $p_0'$. Obviously, for any point $q\in g(c,p_0')$, we always have $p_0' = p_{\gamma_1}(q)$. Let $p_0'' = p_{\gamma_2}(q)$ be the projection of $q$ onto $\gamma_2$. %Clearly, 
%$\big\{d_{\cal M}(q,\gamma_1)-\sigma_{\gamma_1}(p_0')\big\}-\big\{d_{\cal M}(q,\gamma_2)-\sigma_{\gamma_2}(p_0'')\big\}$ is negative for $q\in g(c,p_0')$ sufficiently closing to $p_0'$ or positive if $q$ closes to $p_0''$. 
Hence, identify $q_0\in g(c,p_0')$ such that,
\[
	\mathpzc{d}_{\cal M}(q_0,\gamma_1)-\sigma_{\gamma_1}(p_0') = \mathpzc{d}_{\cal M}(q_0,\gamma_2)-\sigma_{\gamma_2}(p_0'').
\]
Here, we call $q_0$ a warm start.
%By definition, $p_0' = p_{\gamma_1}(q_0)$ and $p_0'' = p_{\gamma_2}(q_0)$. 
Pick a matching pair as follows 
\[
	p_1(0) = p_0',\quad %= p_{\gamma_1}(q_0)
	p_2(0) = p_0''% =p_{\gamma_2}(q_0).
\]
Then we can identify a point $q(0)\in g(p_0',p_0'')$ such that $m_{\gamma_1}(q(0)) = m_{\gamma_2}(q(0))$. Obviously, $q(0)$ satisfies the conditions in Definition \ref{def-class-boundary}.
%(ii) Finding the initial point: 
%Initially, let $\lambda_{\delta}(0) = 1/2$, $\tilde v_1(0)= \dot 	p_1(0) $, $\tilde v_2(0)= \dot 	p_2(0)$. Then, estimate $v(0)$ by
%\begin{align}\label{v_lambda}
%  v_{\delta}(0) = \lambda_{\delta}(0) \tilde v_1(0) + (1-\lambda_{\delta}(0))\tilde v_2(0),
%	%v_{\delta}(0) =& \frac{v_{\delta}(0)}{\left\| v_{\delta}(0)\right\|}
%\end{align}
%and find the projection as
%\begin{align}\label{q_delta}
%	q_{\delta}(0) = \mbox{ {\bf exp}}(q_0, q_0+\delta v_{\delta}(0)).
%\end{align}
%If $q_{\delta}(0)$ satisfies the conditions in Definition \ref{def-class-boundary}, let $q(0) = q_{\delta}(0)$;  otherwise, update $\lambda_\delta(0)$ by \eqref{update} for $t=0$, and re-calculate 
%$v_{\delta}(0)$ in \eqref{v_lambda} and $q_{\delta}(0)$ in \eqref{q_delta}.

Step 2 ({\it Updating the boundary}): Calculating $q(t)$ for $t>0$ from the previous point $q(t-\delta)$ with a small $\delta>0$. Initially, let 
$\tilde v_1(t)= \dot 	p_1(t) $, $\tilde v_2(t)= \dot 	p_2(t)$, and set $\lambda_{\delta}(t)$ as in \eqref{lambda_delta_t}.
Then, estimate $v(t)$ by 
\begin{align} \label{v_lambda_t}
  v_{\delta}(t) = \lambda_{\delta}(t) \tilde v_1(t) + (1-\lambda_{\delta}(t))\tilde v_2(t),
\end{align}
and find the projection as
\begin{align}\label{q_delta_t}
q_{\delta}(t) = \mbox{ {\bf exp}}_{q(t-\delta)}(q(t-\delta)+\delta v_{\delta}(t)).
\end{align}
If $q_{\delta}(0)$ satisfies the conditions in Definition \ref{def-class-boundary}, let $q(t) = q_{\delta}(t)$;  otherwise, update $\lambda_\delta(t)$ by \eqref{update}, and re-calculate 
$v_{\delta}(t)$ in \eqref{v_lambda_t} and $q_{\delta}(t)$ in \eqref{q_delta_t}.

%If the gap
%\[
%	\big\{d_{\cal M}(q_{t+\delta},\gamma_1)-\sigma_{\gamma_1}(p_{t+\delta}')\big\}-\big\{ d_{\cal M}(q_{t+\delta},\gamma_2)-\sigma_{\gamma_2}(p_{t+\delta}'')\big\}
%\]
%is negative, we slightly decrease $\lambda_t$, or increase $\lambda_t$ if the gap is positive. Then we reset $v_t$ in (\ref{def:v_lambda}). Repeat this procedure 
%until the gap is zero or sufficiently small, and set 
%\[
%	q({t+\delta}) = q_{t+\delta},\quad
%	p_1({t+\delta}) = p_{t+\delta}' = p_{\gamma_1}(q_{t+\delta}),\quad
%	p_2({t+\delta}) = p_{t+\delta}'' = p_{\gamma_2}(q_{t+\delta}).
%\]

The algorithm will be executed for a period of time and will produce a sequence of $\{q(t)\}$. The constructed sequence is indeed the principal boundary since we always have that for every $q(t) \in \gamma_c$, $p_1(t) = p_{\gamma_1}(q(t))$, and $p_2(t)= p_{\gamma_2}(q(t))$,
\[
	\mathpzc{d}_{\cal M}(q(t),\gamma_1)-\sigma_{\gamma_1}(p_1(t)) = \mathpzc{d}_{\cal M}(q(t),\gamma_2)-\sigma_{\gamma_2}(p_2(t)).
\]
%As $\gamma_c= \{q(t) : t\in [0,T]\}$, 
Supposing we may discretize the $\gamma_c$ as $\gamma_c=(q(0), \cdots, q(N))$, $q(i) \in \mathcal{M}$. The length of the principal boundary can be numerically approximated
\[
\ell(\gamma_c) \approx \sum_{i=0}^{N-1}d_{\cal{M}}(q(i), q(i+1)).
\]
%where $d$ represents the Riemannian metric on $\mathcal{M}$.

\section{Property of the principal boundary}
\label{Propertyclass}
\subsection{Principal boundary and SVM}
\label{svm-pd}
This section shows that the local segment of the  principal boundary reduces to the boundary given by the support vector machine. We remark here that the SVM boundary used is a manifold extension of the usual SVM, which is essentially a geodesic curve. The same results hold in the context of Euclidean spaces, where $\mathcal{M}$ is a linear subspace of $\mathbb{R}^d$. By making the notion of ``local equivalence'' precise, we provide a measure of distance between the principal boundary, obeying Definition \ref{def-class-boundary} and the SVM boundary. % Suppose that $\gamma_c$ is the classification boundary between the data sets $x_{1,i}$ and $x_{2,i}$, where $\gamma_1$ and $\gamma_2$ are the fitted principal flows, respectively.

%The below shows that the segment of classification boundary, coincides with the segment of SVM, locally.
 
To study the relation, we start with a quantitative description of the segment of $\gamma_c$ on $\mathcal{M}$ by the following proposition. 

\begin{prop}{(Local configuration)} \label{Prop:configuration} Consider a small segment $\gamma_c^*$ of $\gamma_c$ on $\mathcal{M}$. Suppose that 
\begin{itemize}
\item[1.] The segment is discretized as $\gamma_c^*= \{q_1^*, \cdots,q_{M}^*\}$, where $q_j^*  \in \mathcal{M}$.
\item[2.]  Following the notation in Section 3.1, let $\gamma_1^*= \{p_{1,1}^*, \cdots,p_{1,M}^*\}$ be the projections of $\gamma_c^*$ onto $\gamma_1$ and $\mathcal{N}(p_{1,j}^*, h)$ be the set of samples in the $h$-neighborhood of $p_{1,j}^*$.  Clearly, the $M$ local neighborhoods give a configuration of local data points from class $+1$ as 
\[
	\cup_{j=1}^M {\cal N}(p_{1,j}^*, h) = \{x_{1,1},\cdots,x_{1,k_1}\}.
\] 
Similarly, we can define $\{x_{2,1},\cdots,x_{2,k_2}\}$ for the other class $-1$ based on the projections $\gamma_2^*= \{p_{2,1}^*, \cdots,p_{2,M}^*\}$ of $\gamma_c^*$ onto $\gamma_2$.
\end{itemize}
\end{prop}

\begin{figure} 
\centering
          \includegraphics[width=0.45 \linewidth]{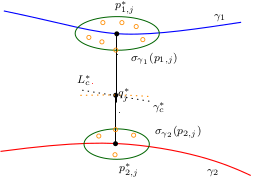}
	 \caption{The principal boundary $\gamma_c^*$ in the local configuration. The covering ellipse ball (in green) with the second radius $\sigma_{\gamma_1}(p_{1,j})$, centered at $p_{1,j}^*$ contains the local points (in ``$\circ$'') for class $+1$; the shortest distance between $q_j^*$ and the corresponding local configuration of class $+1$ is approximated by $d_{\cal M}(p_{1,j}^*,q_j^*) - \sigma_{\gamma_1}(p_{1,j}^*)$ (same for class $-1$). }
					\label{svm-local}
\end{figure}

If $\gamma_c^*$ is small enough, it is approximately a segment of a straight line $L_c^*$, and $q_j^*$s are also approximately located at the line. 
%Let the projections of $q_j^*$ onto $\gamma_1$ and $\gamma_2$ be
%\[
%	\gamma_1^*= \{p_{1,1}^*, \cdots,p_{1,M}^*\},\quad
%	\gamma_2^*= \{p_{2,1}^*, \cdots,p_{2,M}^*\}
%\]
%respectively. 
The SVM on the two classes determines a geodesic curve $L_{\rm svm}$ that separates the two sets
$\gamma_1^*$ and $\gamma_2^*$ such that the margin of $L_{\rm svm}$ 
\[
	m(L_{\rm svm}) = \min\big\{\min_k \mathpzc{d}_{\cal M}(x_{1,k},L_{\rm svm}), \min_k \mathpzc{d}_{\cal M}(x_{2,k},L_{\rm svm})\big\}
\]
is maximized.  For $\gamma_c^*$, we define the margin as 
\[
	m(\gamma_c^*) = \min\big\{\min_k \mathpzc{d}_{\cal M}(x_{1,k},\gamma_c^*), \min_k \mathpzc{d}_{\cal M}(x_{2,k},\gamma_c^*)\big\}.
\] 

To quantify the relation of $m(\gamma_c^*)$  and $m(L_{\rm svm})$,  we will basically need three approximations. We remark here that although a careful approximation of $\gamma_c^*$ to $L_{\rm svm}$ can be bounded under some error assumptions, together with an estimation of $m(\gamma_c^*)\approx m(L_c^*)$, doing so would result in a complicated analysis. To simplify our discussion, we will first sketch an overall review of the local equivalence while highlighting the idea. A refined approximation follows, in terms of Theorem \ref{thm:equivalence}.

Let us consider samples in each neighbor set ${\cal N}(p_{1,j}^*, h)$. Assume that these neighbors are covered by an ellipse ball of the second radius being $\sigma_{\gamma_1}(p_{1,j})$, centered at $p_{1,j}^*$. Since %$d_{\cal M}(p_{1,j}^*,q_j^*) - \sigma_{\gamma_1}(p_{1,j}^*)=m_{\gamma_1}(q_j^*)$,
$m_{\gamma_1}(q_j^*) = d_{\cal M}(p_{1,j}^*,q_j^*) - \sigma_{\gamma_1}(p_{1,j}^*)$ the quantity %with probability 1
\begin{align}\label{approx_1}
%	\min_{x_{1,k}\in {\cal N}(p_{1,j}^*, h)}\mathpzc{d}_{\cal M}(x_{1,k},\gamma_c^*)
%	 %d_{\cal M}(p_{1,j}^*,q_j^*) - \sigma_{\gamma_1}(p_{1,j}^*)
%	 \approx m_{\gamma_1}(q_j^*).
	\min_{x_{1,k}\in {\cal N}(p_{1,j}^*, h)}\mathpzc{d}_{\cal M}(x_{1,k},q_j^*)
	= m_{\gamma_1}(q_j^*).
\end{align}
should approximately hold, up to some degree. Hence,
%From $\min_k \mathpzc{d}_{\cal M}(x_{1,k},\gamma_c^*)
%	= \min_j \min_{x_{1,k}\in {\cal N}(p_{1,j}^*, h)}\mathpzc{d}_{\cal M}(x_{1,k},\gamma_c^*)$, we get that
%Hence, with probability 1, 
\[
	\min_k \mathpzc{d}_{\cal M}(x_{1,k},\gamma_c^*) 
	= \min_j \min_{x_{1,k}\in {\cal N}(p_{1,j}^*, h)}d_{\cal M}(x_{1,k},q_j^*)
	\approx \min_{j}m_{\gamma_1}(q_j^*).
\]
Similarly, we also have that 
$\min_k \mathpzc{d}_{\cal M}(x_{2,k},\gamma_c^*) \approx \min_{j}m_{\gamma_2}(q_j^*)$.  Therefore, we have   
\[
	m(\gamma_c^*) \approx \min_{j}\big\{ m_{\gamma_1}(q_j^*), m_{\gamma_2}(q_j^*)\big\} = \min_{j} m_{\gamma_1,\gamma_2}(q_j^*).
\]

On the other hand, let $q_j'$ be the intersected point of the geodesic curve ${\cal C}(p_{1,j}^*,p_{2,j}^*)$ and the straight line $L_{\rm svm}$. We also have the approximations 
that 
\begin{align}
	&\min_{x_{1,k}\in {\cal N}(p_{1,j}^*,h)} \mathpzc{d}_{\cal M}(x_{1,k},L_{\rm svm})
	\approx d_{\cal M}(p_{1,j}^*,q_j') - \sigma_{\gamma_1}(p_{1,j}^*)
	= m_{\gamma_1}(q_j'), \label{approx_2} \\
	&\min_{x_{2,k}\in {\cal N}(p_{1,j}^*,h)} \mathpzc{d}_{\cal M}(x_{2,k},L_{\rm svm})
	\approx d_{\cal M}(p_{2,j}^*,q_j') - \sigma_{\gamma_2}(p_{2,j}^*)
	= m_{\gamma_2}(q_j'). \label{approx_3} 
\end{align}

It follows that the SVM margin
\begin{align*}
	m(L_{\rm svm}) 
	&= \min_j \min\big\{\min_{x_{1,k}\in {\cal N}(p_{1,j}^*,h)} \mathpzc{d}_{\cal M}(x_{1,k},L_{\rm svm}), %\\ &  \hspace{55bp}		
         \min_{x_{2,k}\in {\cal N}(p_{1,j}^*,h)} \mathpzc{d}_{\cal M}(x_{2,k},L_{\rm svm})\big\}\\
	&\approx \min_j \min\big\{m_{\gamma_1}(q_j'),m_{\gamma_2}(q_j')\big\}
	= \min_j m_{\gamma_1,\gamma_2}(q_j')\\
	&\leq \min_j  \max _{q\in {\cal C}(p_{1,j}^*,p_{2,j}^*)} m_{\gamma_1,\gamma_2}(q)
	=  \min_j m_{\gamma_1,\gamma_2}(q_j^*) = m(\gamma_c^*).
\end{align*}
Let $L_c^*$ be a geodesic curve most close to  $\gamma_c^*$. Then $m(\gamma_c^*)\approx m(L_c^*)$, and 
\[
	m(L_{\rm svm}) \lessapprox m(\gamma_c^*)\approx m(L_c^*)\leq m(L_{\rm svm})
\] 
by definition, which suggests that $L_{\rm svm}$ approximately coincides with $\gamma_c^*$.

%The approximation of $\gamma_c^*$ to $L_{\rm svm}$ can be bounded under error assumptions of %(\ref{approx_1}) and (\ref{approx_2}) (or (\ref{approx_3})), together with an estimation of %$m(\gamma_c^*)\approx m(L_c^*)$, resulting in an complicated analysis. We omit the discussion for %simplicity. 

Obviously, it can be seen that the local equivalence holds if Approximation (\ref{approx_1}), plus (\ref{approx_2}) (or (\ref{approx_3})), are satisfactory. We introduce the following condition to guarantee the approximations up to some quantitative degree of uncertainty. This is done by linking the density of sample points in a local neighbor with a probability measure.  

\begin{condition}{(Covering ellipse ball)} \label{Prop:ellipse} For each $p_{1,j}^*$, consider samples in each neighbor set ${\cal N}(p_{1,j}^*, h)$. We assume that, 
\begin{itemize}
\item[1.] ${\cal N}(p_{1,j}^*, h)$ has $k_{1,j}$ neighbors that are covered by an ellipse ball of the second radius  $\sigma_{\gamma_1}(p_{1,j})$, centered at $p_{1,j}^*$, 
\item[2.] %there are $k_{1,j}$ sample points in the ellipse ball, such that, 
when $k_{1,j} \rightarrow \infty$, with probability of at least $1-o(\frac{1}{\sqrt{k_{1,j}}})$ 
\begin{align*}
	&\Big|\min_{x_{1,k}\in {\cal N}(p_{1,j}^*, h)}\mathpzc{d}_{\cal M}(x_{1,k},\gamma_c^*)-m_{\gamma_1}(q_j^*)\Big| \leq\sqrt{\frac{\log k_{1,j}}{k_{1,j}}};\\
	&\Big|\min_{x_{1,k}\in {\cal N}(p_{1,j}^*, h)}\mathpzc{d}_{\cal M}(x_{1,k},L_{\rm svm})-m_{\gamma_1}(q_j')\Big| \leq\sqrt{\frac{\log k_{1,j}}{k_{1,j}}},
\end{align*}
\item[3.] similar conditions apply to class $-1$.
\end{itemize}
\end{condition}

%{\bf Remark} Condition \ref{Prop:ellipse} essentially 

\begin{thm} \label{thm:equivalence}
Let $\gamma_c^*$ and $L_{\rm svm}$ be the separating boundary between the local samples of two classes, derived by the principal boundary and SVM, respectively. Let $k = \min_{i,j}k_{i,j}$ where $i=1,2$ and $j=1, \cdots, M$. Given Proposition \ref{Prop:configuration} and Condition \ref{Prop:ellipse}, $\gamma_c^*$ and $L_{\rm svm}$ are equivalent such that $m(\gamma_c^*) =m(L_{\rm svm})$, with probability of at least $1-o(\frac{1}{\sqrt{k}})$, for $k \rightarrow \infty$.
\end{thm}

Theorem \ref{thm:equivalence} gives an equivalence of  $\gamma_c*$ and $L_{\rm svm}$ on the curved manifold. The proof of Theorem \ref{thm:equivalence} is given in Appendix A \cite{Boundarysupp}. Although we have potentially linked $L_{\rm svm}$ with $\gamma_c*$ with an interest of interpreting them locally, it does not necessarily mean that Theorem \ref{thm:equivalence} is only valid when the locality is infinitesimal.  Instead, it is governed by the spacings of the segment $\gamma_c^*$. In fact, when the locality of $\gamma_c^*$ is not infinitesimal, the equivalence still holds between $L_c^*$ and $L_{\rm svm}$, provided that $L_c^*$ is close to $\gamma_c*$. In case of a flat manifold, the results still hold, where both $L_{\rm svm}$ and $\gamma_c*$ are reduced to straight lines, and $\gamma_c*$ is a curve close to them.

\subsection{Random principal boundary}
\label{random-pd}
\bed (Random distribution along a curve)\label{random-pts}
Let a unit-speed curve $\gamma^0= \{q(t): t \in [0,T]\}$ be the population curve.  For simplicity, assume $\mbox{dim}(\gamma^0)=1$. The random points along $\gamma^0$ are defined by the following procedure: 
for each $q(t) \in \gamma^0$, we assume that, 
\begin{itemize}
\item[1.] the normal plane $N^{\bot}_{q(t)} \gamma^0$ exists, 
\item[2.] $q(t) \sim g$, where $g$ is a uniform distribution on $\gamma^0$,
\item[3.] the conditional distribution $f$ is well defined on the intersection of the normal plane $N^{\bot}_{q(t)} \gamma^0$ and the tangent plane $T_{q(t)}\cal{M}$, with the mean $E(f)=q(t)$ and $\mbox{Cov}(f)=\Sigma_{q(t)}$. For simplicity, let $f \sim N(q(t), \Sigma_{q(t)})$.
\end{itemize}
\eed

%%%%%%%%%%%%%%%
{\bf Remark 1:}  Condition 3 allows $\gamma^0$ to be extendable to a sub-manifold ($\mbox{dim}(\gamma^0) \leq m-1$). The $\Sigma_{q(t)}$ defined is a $d\times d$ matrix %which is 
of rank $m-\dim\gamma^0$, i.e., $\Sigma_{q(t)}$ is not full rank. Thus, the $f$ defined is degenerate and does not have a density function with respect to the $d$-dimensional Lebesgue measure in $\mathbb{R}^d$. In the proof, to define the densities and avoid complications, we let $\Sigma_{q(t)}$ be a full rank $(m-\dim(\gamma^0)) \times (m-\dim(\gamma^0))$ matrix in the local coordinates.

{\bf Remark 2:} The intersection of $N^{\bot}_{q(t)}\gamma^0$ and $T_{q(t)}\cal{M}$ is of dimension $m-\mbox{dim}(\gamma^0)$ in $\mathbb{R}^d$ ($\dim(\gamma^0)=1$ for principal flow).
 We assume that $\gamma^0$ contains the origin, $o$, then $T_o\mathcal{M}, N^{\bot}_o\mathcal{M}, T_o\gamma^0$ and $N^{\bot}_o\gamma^0$ are subspaces with
%\begin{align*}
$\dim(T_o\mathcal{M}) = m, \dim(N^{\bot}_o\mathcal{M}) = d-m,  %\\
\dim(T_o\gamma^0) = \mbox{dim}(\gamma^0)$, and $\dim(N^{\bot}_o\gamma^0) = d-\mbox{dim}(\gamma^0)$, respectively.
%\end{align*}
Note that $(T_o \mathcal{M} \cap N^{\bot}_o\gamma^0) \oplus N^{\bot}_o\mathcal{M} = N^{\bot}_o\gamma^0$, therefore 
%\begin{align*}
$\dim(T_o \mathcal{M} \cap N^{\bot}_o\gamma^0) = \dim(N^{\bot}_o\gamma^0) - \dim( N^{\bot}_o\mathcal{M})=(d-\mbox{dim}(\gamma^0))-(d-m)=m-\mbox{dim}(\gamma^0)$.\\%\\
%$&=(d-k)-(d-m)\\
%$&=m-k
%\end{align*}}
%%%%%%%%%%%%%%%

\begin{figure}[ht]
\centering
\includegraphics[width=2 in]{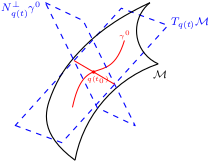}
\caption{Sketch of random distribution along a curve.}
\label{random-setup}
\end{figure}

\begin{definition}{(Continuous principal flow for a random distribution)} \label{Prop:PF-cont} Consider the distribution of Definition \ref{random-pts} in each neighbor set $D ={\cal N}(q(t), h)$, define the continuous principal flow %$\gamma^{0'}_{f}=\gamma^{0'}=\{q'(t): t \in [0,T]\}$ 
$\tilde \gamma^{0}_{f}=\tilde \gamma^{0}=\{\tilde q(t): t \in [0,T]\}$
such that, for each %$q'(t) \in \gamma^{0'}$, $$q'(t)=E_{g, f}(q)$$
$\tilde q(t) \in \tilde\gamma^{0}$, 
$$\tilde q(t)=E_{g, f}(q),$$
where $q \in D$.
\end{definition}

{\bf Remark:} This definition is for the continuous principal flow with respect to distributions $f$ and $g$. Although $\mbox{dim}(\gamma^0)$ is implicitly defined to be 1, Definition \ref{random-pd} and \ref{Prop:PF-cont} still holds for any $\mbox{dim}(\gamma^0) \leq m-1$.

\begin{lemma}{(Convergence of continuous principal flow)} \label{lm:convergence-pf} For a fixed $h$ and $t$, assume that there exists a one-to-one correspondence between $q(t) \in \gamma^0$ and $\tilde q(t)\in \tilde\gamma^{0}$, where $\gamma^0$ and $\tilde\gamma^{0}$ are defined in Definition \ref{random-pts} and \ref{Prop:PF-cont}. Then, if we choose $\sigma_{q(t)}$ such that $f$ is concentrated inside the neighborhood set $D$ with probability at least $1-\varepsilon$, then $\|\tilde q(t)-q(t)\|_{2}=O(\varepsilon)$ for each $t$.
\end{lemma}

{\bf Remark:} The proof in Appendix A \cite{Boundarysupp} is given for a 2 and 3-dimensional manifold. The same proof can be extended up to a $(m-1)$-dimensional manifold. %Here we choose $h$ small enough such that the curve $q(t)$ is approximately equal to its tangent in the tangent plane. Then the $z$ component is linearly dependent on $x$ and $y$, i.e. $z=ax+by+c$ for some $a,b$ and $c$. We see that for $q(t)=(x_0,y_0,z_0)$, $\tilde q(t)=(x_0,y_0+O(\varepsilon),z_0+O(\varepsilon))$, then $\|\tilde q(t)-q(t)\|_{2}=O(\varepsilon)$.

Let $f$ be the function that finds the $q_c(t) \in \gamma_c$ given $q_1(t)\in \gamma_1$ and $q_2(t) \in\gamma_2$ and assume that $f$ is Lipschitz continuous with constant $L$. Assume that $\|\tilde q_1(t)-q_1(t)\|_{2}=O(\varepsilon_1)$ with probability of at least $1-\varepsilon_1$ and $\|\tilde q_2(t)-q_2(t)\|_{2}=O(\varepsilon_2)$ with probability of at least $1-\varepsilon_2$. Then, with probability of at least $1-\max(\varepsilon_1,\varepsilon_2)$, $\|\tilde q_c(t)-q_c(t)\|_{2}=L (O(\varepsilon_1)+O(\varepsilon_2))$ for each $t$.

\begin{lemma}{(Convergence of continuous principal boundary)} \label{lm:convergence-pb} 
%Let $\gamma_1$ and $\gamma_2$ be the two population curves. Let $\tilde\gamma_1$ and $\tilde\gamma_2$ be the continuous principal flows of $\gamma_1$ and $\gamma_2$, respectively. Then, under the same conditions in Definition  \ref{random-pts} and \ref{Prop:PF-cont}, the continuous boundary $\tilde\gamma_c$ of $\tilde\gamma_1$ and $\tilde\gamma_2$ is point consistent to $\gamma_c$, the population boundary of $\gamma_1$ and $\gamma_2$. That is, let $f$ be the function finds the $q_c(t) \in \gamma_c$ given $q_1(t)\in \gamma_1$ and $q_2(t) \in\gamma_2$ and assume that $f$ is Lipschitz continuous with constant $L$. Assume that $\|\tilde q_1(t)-q_1(t)\|_{2}=O(\varepsilon_1)$ with probability of at least $1-\varepsilon_1$ and $\|\tilde q_2(t)-q_2(t)\|_{2}=O(\varepsilon_2)$ with probability of at least $1-\varepsilon_2$. Then, with probability of at least $1-\max(\varepsilon_1,\varepsilon_2)$, $\|\tilde q_c(t)-q_c(t)\|_{2}=L (O(\varepsilon_1)+O(\varepsilon_2))$ for each $t$.
Let $\tilde\gamma_1$ and $\tilde\gamma_2$ be the continuous principal flows of two population curves $\gamma_1$ and $\gamma_2$, respectively. 
Then, under the same conditions in Definition  \ref{random-pts} and \ref{Prop:PF-cont}, the continuous boundary $\tilde\gamma_c$ of $\tilde\gamma_1$ and $\tilde\gamma_2$ is a point consistent to the population boundary $\gamma_c$ of $\gamma_1$ and $\gamma_2$. 
\end{lemma}

{\bf Remark:}  To fulfill the classification purpose, we need $\mbox{dim}(\gamma^0)=m-1$ for the principal boundary. Without loss of generality, we only prove (see Appendix A \cite{Boundarysupp}) the case of $\gamma_c$ being the principal boundary ($m=2$) with zero margins to $\gamma_1$ and $\gamma_2$. The proof to non-zero margins would be the  same, so we omit it.

\begin{figure}[ht]
\centering
\begin{subfigure}[b]{0.3 \textwidth}
\includegraphics[width=1.5 in]{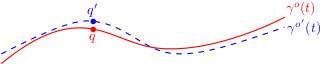}
\caption{}
\end{subfigure}
\hspace{0.3 in}
\begin{subfigure}[b]{0.3\textwidth}
\includegraphics[width=1.5in]{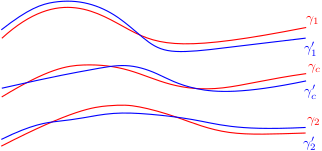}
\caption{}
\end{subfigure}
\caption{Consistency of the principal boundary under random distribution: (a) Population curve (red) and continuous principal flow (blue). (b) Population curves ($\gamma_1, \gamma_2$) and population boundary ($\gamma_c$), continuous principal flows ($\tilde\gamma_1, \tilde\gamma_2$) and continuous boundary ($\tilde\gamma_c$).}
\label{bplot}
\end{figure}

%\newpage

\section{Simulation}\label{sim}
\subsection{Principal boundary and SVM}\label{sim1}
To illustrate that the principal boundary dependent on the configuration of the data sets, we first generate two sets of noisy data on $S^2 \subset \mathbb{R}^3$. We claim here that choosing the sphere as a test manifold is done only for simplicity, as it stands for the most common manifold that one can work with and compare with other methods. By tuning the vector field along the  principal flow of the two data sets, we visualize the evolution of the principal boundary.  We remark here that the principal flows are the modified principal flows in \eqref{mflow} for the labeled data.
 Figure \ref{bplot} shows the labeled data (Figure \ref{bplot}(a)), the super-imposed principal flows (Figure \ref{bplot}(b)), and the principal boundary (Figure \ref{bplot}(c-d)). In the step of initializing the boundary, the initial point (in red) has been obtained from the warm start (in blue); both of them are labeled in Figure \ref{bplot}(c).  In Figure \ref{bplot}(b), the two green curves are the estimated deviation of the principal flows, providing a measure of the margin of the flow. The boundary enables itself to bend wherever the vector field of the two flows changes. This can be seen from the vectors (in red/black arrows) of Figure \ref{bplot}(d), where they are optimized towards a direction of achieving the maximum margin between the two flows, a criterion introduced in the step of updating the boundary. 
\begin{figure}[ht]
\centering
\begin{subfigure}[b]{0.24\textwidth}
\includegraphics[width=1.5in]{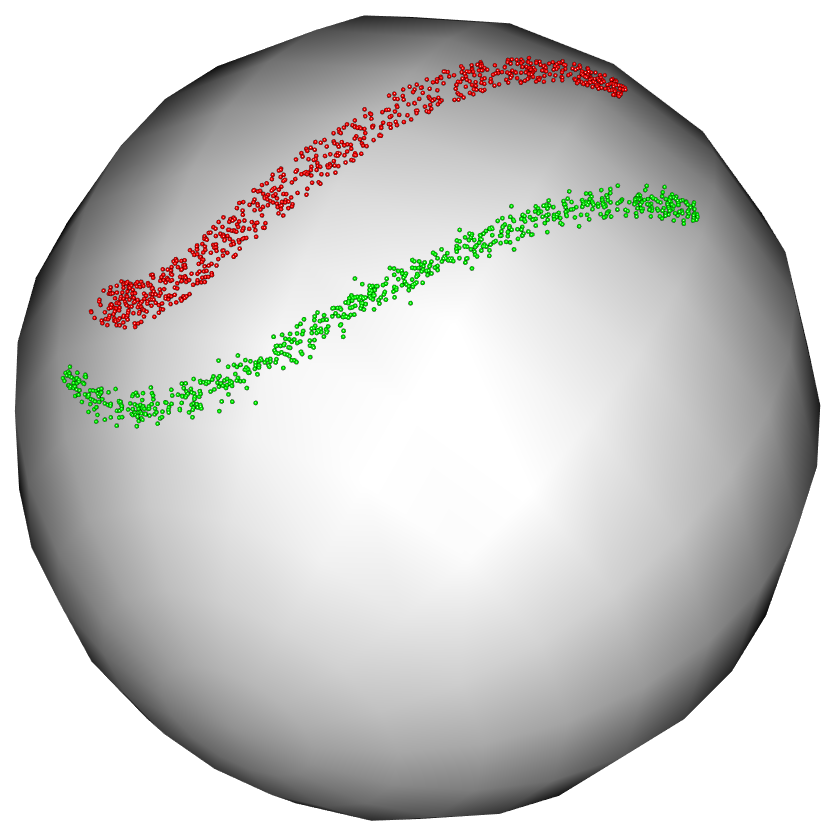}
\caption{}
\end{subfigure}
\hspace{0.45 in}
\begin{subfigure}[b]{0.24\textwidth}
\includegraphics[width=1.5in]{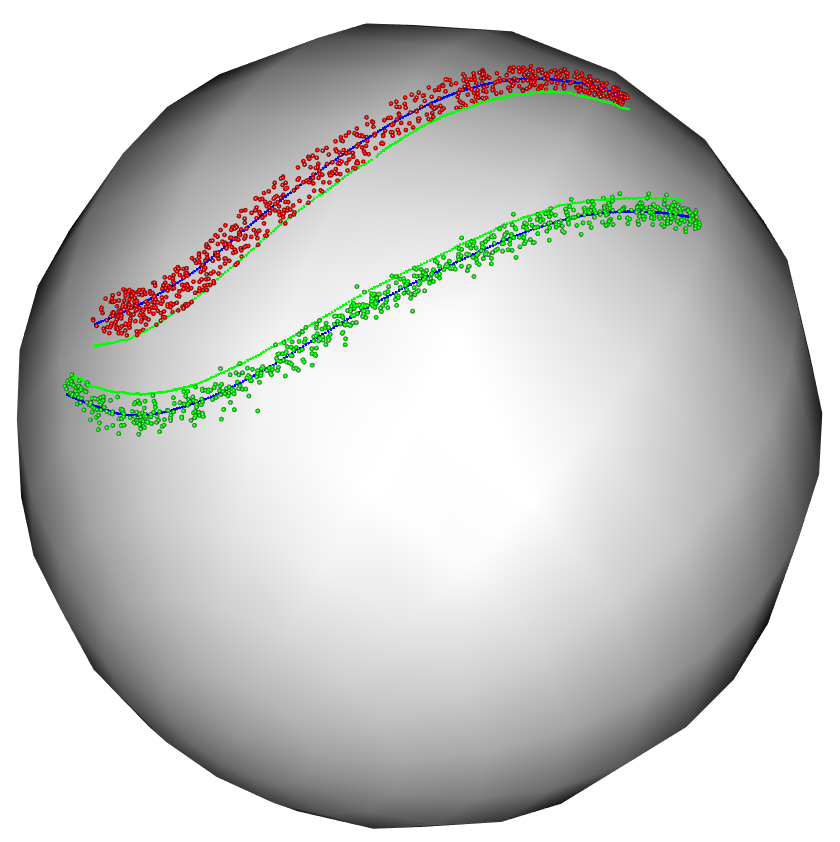}
\caption{}
\end{subfigure}\\
\begin{subfigure}[b]{0.24\textwidth}
\includegraphics[width=1.5in]{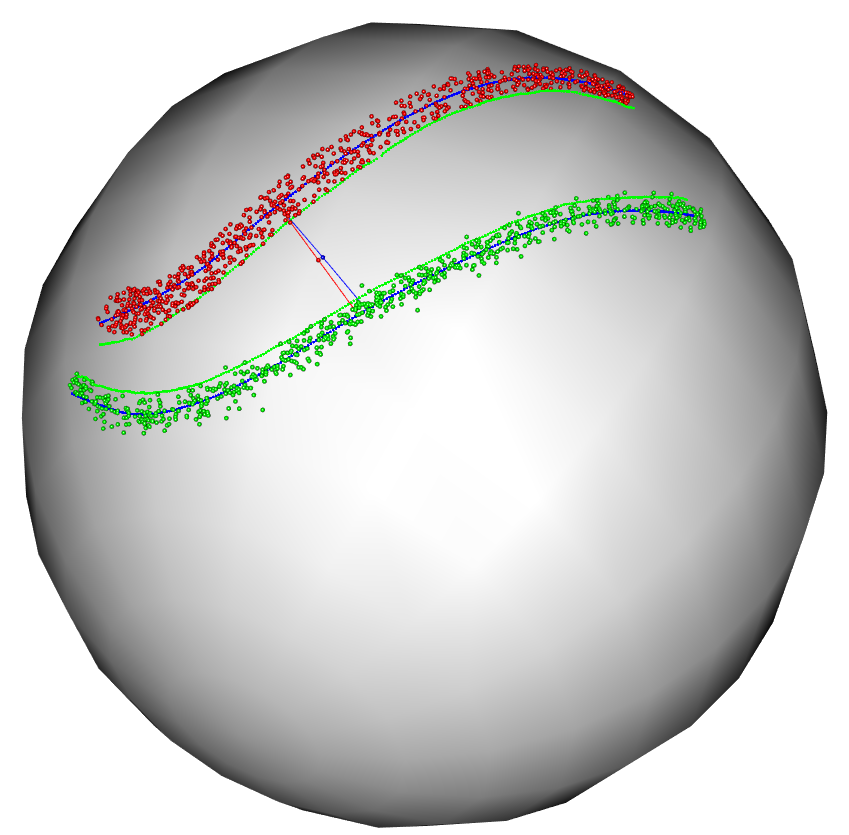}
\caption{}
\end{subfigure}
\hspace{0.45 in}
\begin{subfigure}[b]{0.24\textwidth}
\includegraphics[width=1.5in]{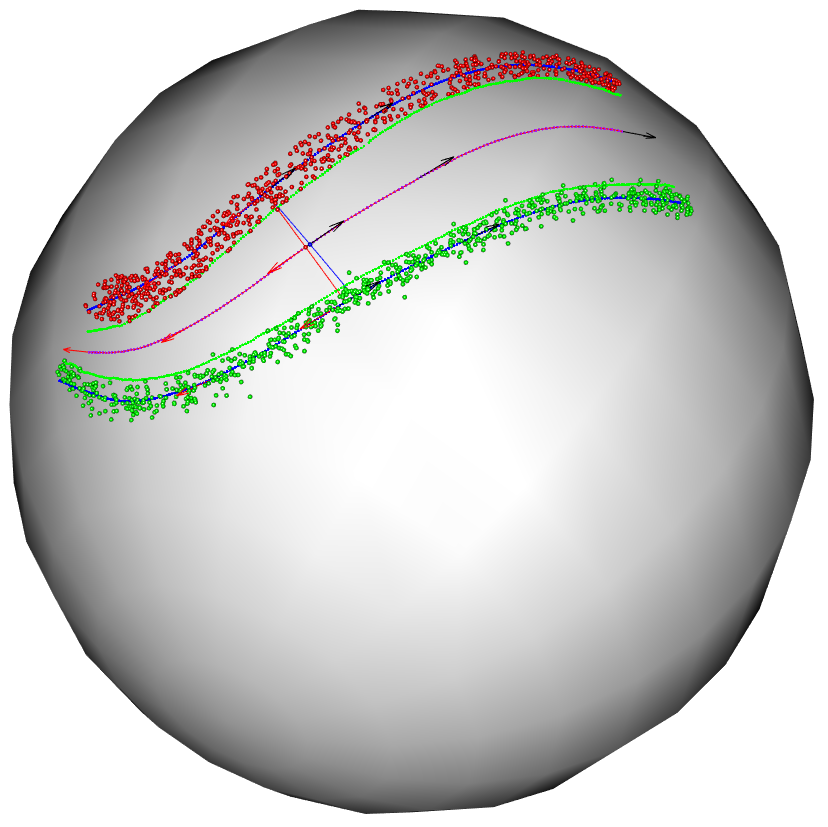}
\caption{}
\end{subfigure}
\caption{Plots of the principal boundary: (a) Data points. (b) Principal flow with estimated margin. %Modified principal flow fitted to the data with an estimated margin. 
(c) Initialization of the boundary. (d) Full path of the principal boundary.}
\label{bplot}
\end{figure}

\begin{figure}[t]
\centering
\begin{subfigure}[b]{0.24\textwidth}
\includegraphics[width=1.5in]{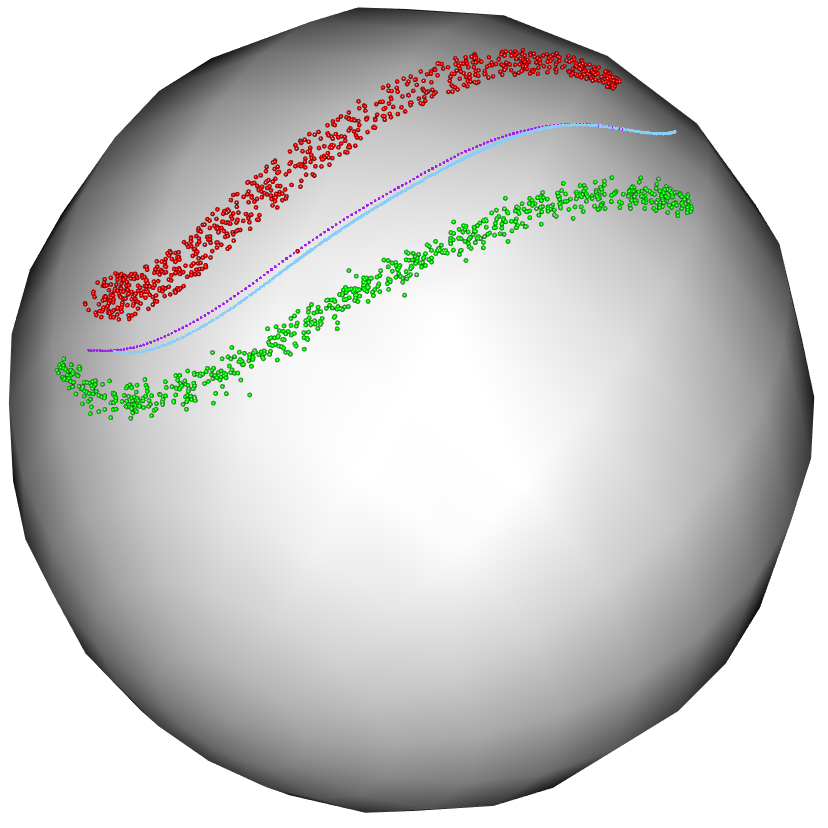}
\caption{}
\end{subfigure}
\hspace{0.45 in}
%\begin{subfigure}[b]{0.22\textwidth}
%\includegraphics[width=1.5in]{PDF/bplots-redo/svm5-redo}
%\caption{}
%\end{subfigure}\\
\begin{subfigure}[b]{0.24\textwidth}
\includegraphics[width=1.5in]{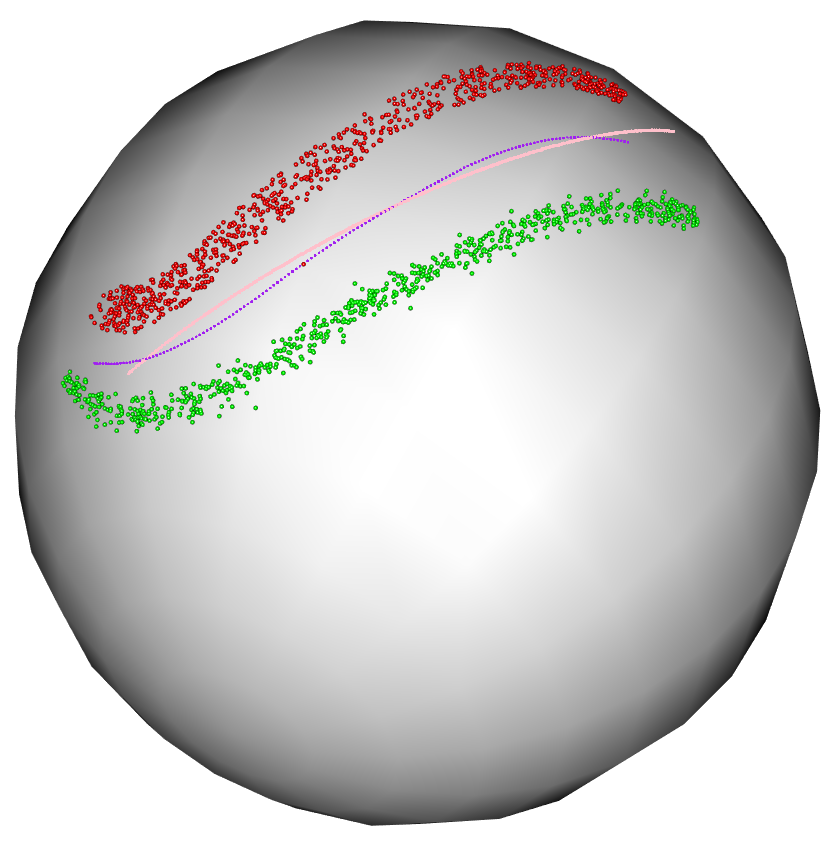}
\caption{}
\end{subfigure}\\
\begin{subfigure}[b]{0.24\textwidth}
\includegraphics[width=1.5in]{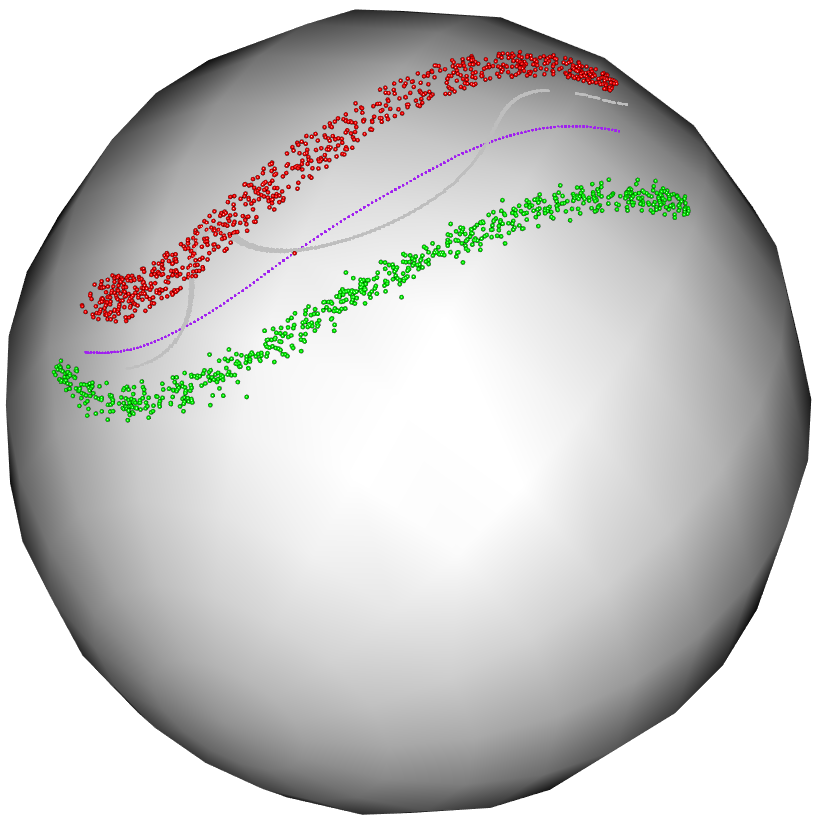}
\caption{}
\end{subfigure}
\hspace{0.45in}
\begin{subfigure}[b]{0.24\textwidth}
\includegraphics[width=1.5in]{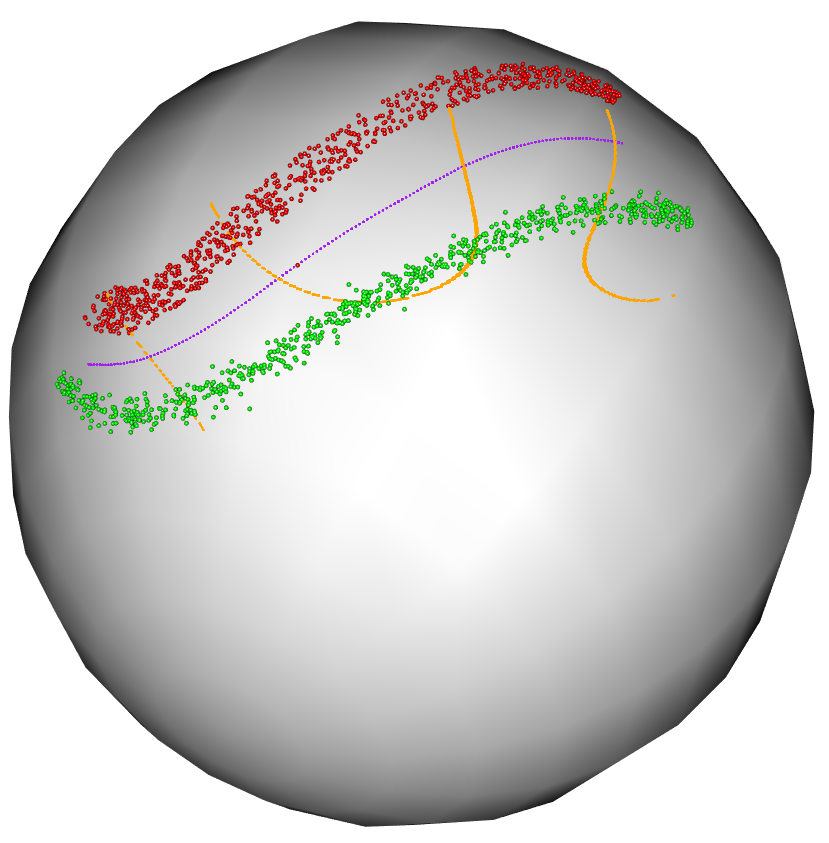}
\caption{}
\end{subfigure}
\caption{
%SVM boundary and principal boundary on the unit sphere: {\color{red}(a) SVM boundary (RBF: $K(u,v)=\mbox{exp}(-\gamma(\|u-v\|^2))$) and principal boundary. (b) SVM boundary (linear: $K(u,v)=u'v$) and principal boundary. (c) SVM boundary (polynomial: $K(u,v)=\left(\gamma u'v+c_0\right)^{\mbox{{\small df}}}$) and principal boundary. (d) SVM boundary (sigmoid: $K(u,v)=\mbox{tanh}\left(\gamma u'v+c_0\right)$) and principal boundary. All kernel functions except the linear function have been used with default parameters: bandwidth $\gamma=1/4$ and Lagrange multiplier  $C=1$. For polynomial, the degree parameter $\mbox{df}=3$.} For polynomial and sigmoid, the coefficient parameter $c_0=0$.
Principal boundary and SVM boundary with different kernel functions: (a) RBF: $K(u,v)=\mbox{exp}(-\gamma(\|u-v\|^2))$. (b) Linear: $K(u,v)=u'v$. (c) Polynomial: $K(u,v)=\left(\gamma u'v+c_0\right)^{\mbox{{\small df}}}$. (d) Sigmoid: $K(u,v)=\mbox{tanh}\left(\gamma u'v+c_0\right)$.  All the parameters are set as: bandwidth $\gamma=1/4$, Lagrange multiplier  $C=1$, $\mbox{df}=3$, and the coefficient parameter $c_0=0$.
}
\label{svmbound}
\end{figure}

We also contrast the principal boundary with the boundary by the SVM classifier on the same set of data. 
Among all of the experiments, the principal boundary (in purple) goes through the initial point (in red), maintaining an equal margin between the two labeled data sets.
We generate the SVM boundary by performing an SVM classifier with a given kernel function, separating the data in the Euclidean space before mapping it back onto the manifold. 
%By avoiding  explicit mapping in SVM, four kernels (radial basis function (RBF), linear, polynomial and sigmoid) have been used with the SVM in training the classification machine. Their performances have been compared with the principal boundary and the result is reported in Figure \ref{svmbound}.  
Four kernels with suitable parameters have been used in the SVM for training the classification machine: radial basis function (RBF) $K(u,v)=\mbox{exp}(-\gamma(\|u-v\|^2))$, linear kernel $K(u,v)=u'v$, polynomial kernel $K(u,v)=\left(\gamma u'v+c_0\right)^{\mbox{{\small df}}}$,
and sigmoid kernel $K(u,v)=\mbox{tanh}\left(\gamma u'v+c_0\right)$. Their performances are compared with the principal boundary and the result is reported in Figure \ref{svmbound}. 
%Among all of the experiments, the principal boundary (in purple) goes through the initial point (in red), remaining an equal margin between the two labelled data sets. 
The SVM finds the boundary quite differently in all four cases: the best boundary found by the SVM is the one with RBF (in light blue), due to the fact that it is the closest one to the principal boundary, with the exception of part on the left (Figure \ref{svmbound}(a)); the linear kernel (in pink)  does a decent job in the middle part of the data but clearly could not handle the curvature on the two ends (Figure \ref{svmbound}(b)); both the polynomial (in grey) and sigmoid (in orange) kernels fail to achieve a reasonable boundary (Figure \ref{svmbound} (c), (d)) on the manifold. More experiments on the SVM boundary with varying parameters under different kernels can be found in Appendix B \cite{Boundarysupp}. This example shows the impact of the resulting SVM boundary from choosing different kernel functions in SVM. If the choice of kernel is done in advance, it can be seen that SVM performs relatively well (i.e., in this case, when the RBF kernel is chosen) compared to the principal boundary. However, very often, one is not given an indication of which kernel to use when performing the classification using the SVM. In this sense, the principal boundary method could be a wise choice when the data clearly has a structure.

\begin{figure}[ht]
\centering
\includegraphics[width=2in]{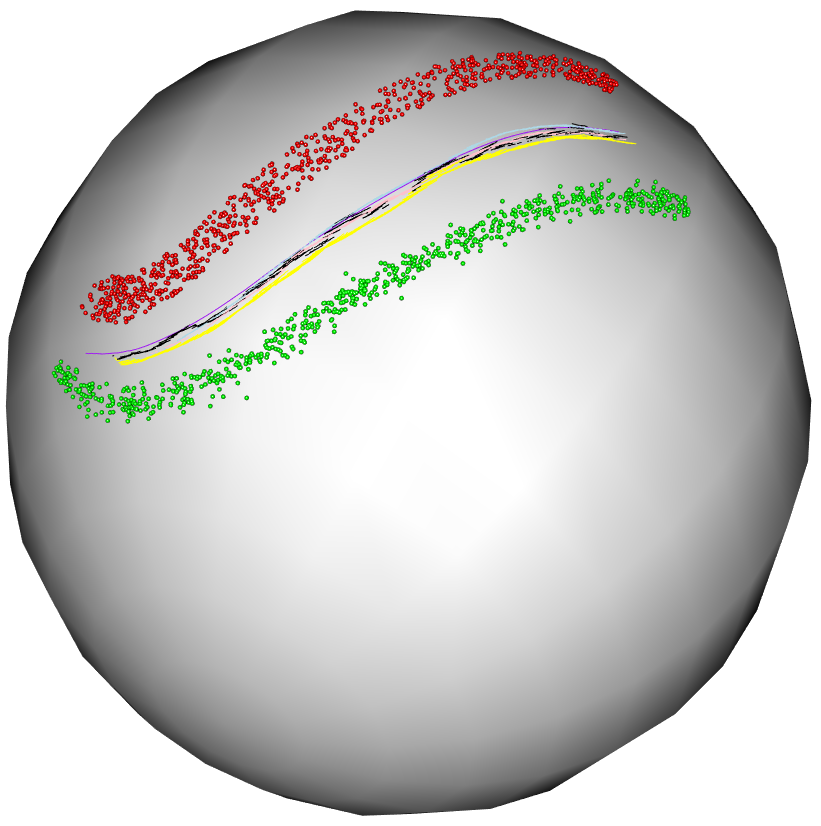}
\caption{Performance of piecewise SVM boundary and the principal boundary. The piecewise SVM boundary with different $h$ are plotted in black ($h=.05$), yellow ($h=.1$), pink ($h=.15$) and light-blue ($h=.2$). The principal boundary (purple) is also superimposed.}
\label{piecewise}
\end{figure}

Though the SVM boundary above seems to be a reasonable means of understanding the principal boundary, a refined SVM boundary has also been investigated. In line with the setting of local configuration \ref{Prop:configuration}, we have obtained a piecewise SVM boundary for this same set of data. This boundary is constructed by performing the SVM process on each pair of the corresponding neighborhoods from the two classes. In this sense, the refined boundary is a local version of the previous SVM boundary. As showed in the proof of Theorem \ref{thm:equivalence}, the local SVM is preformed on each paired neighborhood--essentially, it is a geodesic segment or rough line segment; there is no significant difference in choosing the type of kernel. Figure \ref{piecewise} shows the piecewise SVM boundary at each locality parameter $h \in (.05, .1, .15, .2)$ via linear kernel. It is expected that the piecewise SVM boundary would not necessarily produce a smooth boundary, as we observe that when $h$ increases, the discontinuity of the boundary improves, with a common trend that all SVM segments as a whole are aligned closer to the principal boundary. Although this does not necessarily suggest the existence of an optimal $h$ here, we do see that this behavior matches well with that of the aforementioned Theorem \ref{thm:equivalence}.

\subsection{Principal boundary in an overlap case}\label{sim2}
\begin{figure}[ht]
\centering
\includegraphics[width=2.5in]{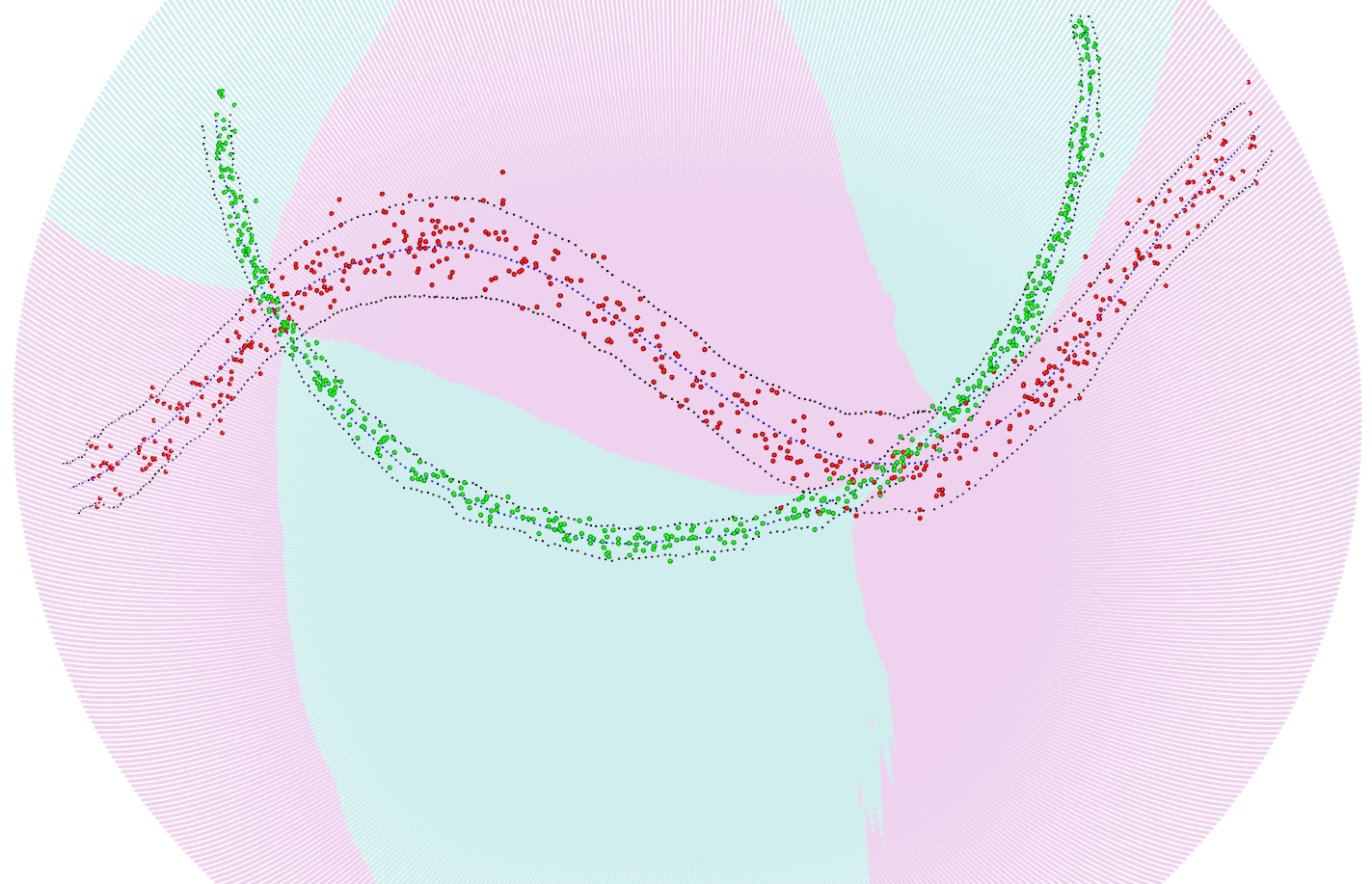}
\caption{C-S data: C-type points (in green) and S-type points (in red). The principal flows and the two-sided margin flows are plotted in black and blue curves, respectively. The region on $S^2$ are labeled by pink and light green according to the classification rule.}
\label{cs}
\end{figure}

To assess the classification performance, we continue to generate  data on $S^2 \subset \mathbb{R}^3$ that consists of two classes. The two classes overlap in some regions. There are 500 points in each class. Given two sets, C-type $C_1$ (in green) and S-type $C_2$ (in red) in Figure \ref{cs}, where the points in $C_1$ are  labeled as $+1$ and the points in $C_2$ are labeled as $-1$. The goal is to assign the labels of points in the regions near $C_1$ and $C_2$. 

We find the respective principal flows (in blue) $\gamma_1$ and $\gamma_2$, and the respective margin flows (in black) $\tilde\gamma_1$ and $\tilde\gamma_2$ ($\tilde\gamma_i$ is two-sided along $\gamma_i$, $i\in 1,2$). For a point $p$ on ${\cal M}$, we define the distance $\mathpzc{d}_{\cal M}(p,C_i)$ of a point $p$ to a set $C_i, i=1,2$ as $$\mathpzc{d}_{\cal M}(p,C_i)=m_{\gamma_i}(p).$$ %[\mathpzc{d}(P,\gamma_i)-margin(P,F_i)]_{+}$$
Note that $\mathpzc{d}_{\cal M}(p,C_i)=0$ implies that the point $p$ falls between the two margins of the set $C_i$, which can be seen to mean that $p$ is inside the set $C_i$. To classify $p$, we use the classification rule as follows:
		\begin{itemize}
			\item[1] If $\mathpzc{d}_{\cal M}(p,C_1)<\mathpzc{d}_{\cal M}(p,C_2)$, $p$ is classified to $C_1$; otherwise $p$ is classified to $C_2$;
			\item[2] If $\mathpzc{d}_{\cal M}(p,C_1)=\mathpzc{d}_{\cal M}(p,C_2) \neq 0$, $p$ is on the classification boundary;
			\item[3] If $\mathpzc{d}_{\cal M}(p,C_1)=\mathpzc{d}_{\cal M}(p,C_2)=0$, $p$ falls into the overlapping area of sets $C_1$ and $C_2$.
		\end{itemize}	

Where $p$ falls into the overlapping area of sets $C_1$ and $C_2$, we will use the relative gap, the ratio of the geodesic distance between $p$ and $\gamma_i$ to the corresponding margin in $C_i, i=1,2$:
\begin{align}\label{adher} R(p,C_i)=\mathpzc{d}_{\cal M}(p,\gamma_i)/\sigma_{\gamma_i}(p).\end{align}
Note that $R(p,C_i)$ describes the adherences of $p$ to $C_i$. The larger this ratio is, the less the adherence of $p$ to $C_i$ is. This gives the following classification rule for the overlapping area: $p$ is classified to $C_1$ when $R(p,C_1)<R(p,C_2)$.\\

\noindent {\bf Remark}: In general, one can set the ratio as $R(p,C_i)=\mathpzc{d}_{\cal M}(p,\gamma_i)^{\alpha}/\sigma_{\gamma_i}(p)^{\beta}$. \eqref{adher} is the case where $\alpha=\beta=1$. In practice, one can choose to tune $\alpha$ and $\beta$ within a reasonable range.
\begin{table}[ht]
\centering % used for centering table
\begin{tabular}{c c c c c c} % centered columns (4 columns)
\hline\hline %inserts double horizontal lines
  &  principal boundary  & SVM (R, L, P, S) & random forest  & $K$-means \\ [0.5ex] % inserts table
\hline % inserts single horizontal line
Error &   .036  & .05, .423, .075, .45 &  .039 & .479&   \\ % inserting body of the table
\hline %inserts single line
\end{tabular}
\caption{Comparison of principal boundary (Column 2) with SVM, random forest and $K$-means (Column 4, 5, 6) on classification errors. SVM (R, L, P, S) refers to SVM using RBF, linear, polynomial and sigmoid kernel.}
\label{cscomp}
\end{table}

We compare the performance of the principal boundary with SVM, random forest and $K$-means. We report an error rate over the C-S data in Table \ref{cscomp}. For the principal boundary, the best error rate is calculated by computing the principal flows over $h_1 \in (.1, .25)$ with an increment $.01$ and $h_2 \in (.05, .15)$ with an increment $.01$. The best errors of SVM with different kernels have been computed over the same range of parameters used in Section \ref{sim2}. We also include the best error rates from random forest and $K$-means. The supplementary materials contain more details on the implementation procedures for all the methods. In this example, the principal boundary has a much better error rate, even though the dimensionality of the ambient space is not large.

\section{Analysis of Seismic Events}
\label{realdata}
We consider a dataset of seismic events involving earthquakes and volcanoes. Volcanoes and earthquakes both result from the movement of tectonic plates and are most likely to occur either on or near plate boundaries. In our study, the volcano dataset contains information about major volcano activity in 2001,\footnote{It can be found at \textit{http://legacy.jefferson.kctcs.edu/techcenter/gis\%20data/World/Zip/VOLCANO.zip}.} The earthquake data contains information about significant earthquakes (measured by damage caused, i.e., losses of approximately 1 million US dollars or more, 10 or more deaths, a magnitude of 7.5 or greater, Modified Mercalli Intensity X or greater, or the earthquake generated a tsunami) that happened between 1900 and 2018.\footnote{It is available from the website of National Oceanic and Atmospheric Administration (doi:10.7289/V5TD9V7K).} 

%%%%%%%%%%%%%%%%%%%%%%%%%%%%%%%%%%%%%%%%%%%%%%%%%%%%%%%%%%%%%%%%%%%
%%%%%%%%%%%%%%%%%%%%%%%%%%%%%%%%%%%%%%%%%%%%%%%%%%%%%%%%%%%%%%%%%
%\subsection{Offshore earthquakes and volcanoes of eastern Japan}

%Figure \ref{data_offshore} shows the major volcano data ($135\degree - 150\degree$ E, $30\degree - 56\degree$ N) in 2001 and significant offshore earthquake data ($140\degree - 146\degree$ E, $30\degree - 44\degree$ N) between 1900 and 2018 that occurred around the region of eastern Japan.

%\begin{figure}[h]
%\centering
%\includegraphics[width=2.5 in]{PDF_Data/japanflatplot_offshore.pdf}
%\caption{Seismic events data of significant offshore earthquakes with magnitude 7.5 or greater (in blue) between 1900 and 2018 and major volcanoes (in red) in 2001 on a flat world atlas.}\label{data_offshore}
%\end{figure}

Besides intense volcano activity, Japan is one of the places in the world that is most affected by significant earthquakes. We extract the seismic events that occurred on the eastern side of Japan and consider the significant offshore earthquakes and major volcanoes of eastern Japan. The principal boundary has been applied to the volcano and offshore earthquake dataset, where the sample size of significant offshore earthquake data $n_\text{eq}=96$ and the sample size of major volcano data $n_\text{vol}=71$, with two varying sequences of scale parameters $h_\text{vol}$ and $h_\text{eq}$. Table \ref{error_offshore} shows the number of misclassified pairs $(m_\text{vol},m_\text{eq})$ in the offshore earthquake and volcano data from varying $h_\text{vol}$ and $h_\text{eq}$. We calculate the overall error rate ($(m_\text{vol}+m_\text{eq})/(n_\text{vol}+n_\text{eq})$) and find that the overall error rate achieves its minimum 0.0120, where there are 0 misclassified points for the volcano data and 2 misclassified points for the offshore earthquake data, among all the cases. The supplementary material contains a similar analysis for the same volcano data and significant earthquakes with magnitude 6 or greater within the same geographic area.

\begin{table}[h]
%\scriptsize
\centering
\begin{tabular}{ c c c  c c c  c c c  c c c }
\hline\hline
$h_\text{vol} \backslash h_\text{eq}$ & 0.020 & 0.025 & 0.030 & 0.035  & 0.040  & 0.045  & 0.050 \\\hline
0.0350 &   (0,2)  &  (0,2) &   (0,3)  &  (1,3)  &  (0,3)  &  (0,3)  &  (0,3) \\
0.0400   &   (0,3) &   (0,3)  &  (0,3) &   (1,3)  &  (0,3)  &  (0,3)  &  (0,3)\\
0.0450 & (0,4)  &  (0,4) &   (0,5)  &  (0,4)  &  (1,4)  &  (1,6) &   (1,6) \\
0.0500  & (0,4)  &  (0,4) &   (0,6)  &  (0,4)  &  (1,5) &   (2,6)  &  (2,6) \\
0.0550   &  (0,4) &   (1,5) &   (0,8)  &  (0,6) &   (1,6)  &  (3,6)  &  (3,6)\\
0.0600 & (0,5)  &  (1,5)  &  (0,7)  &  (0,6)  &  (0,6)  &  (2,6) &   (2,6)\\\hline\hline
\end{tabular}
\caption{The number of misclassified pairs $(m_\text{vol},m_\text{eq})$ in the offshore earthquake and volcano data from varying scale parameters $h_\text{vol}$ and $h_\text{eq}$.}\label{error_offshore}
\end{table}

\begin{figure}[t]
\centering
\begin{subfigure}[b]{0.3\textwidth}
\includegraphics[width=2 in]{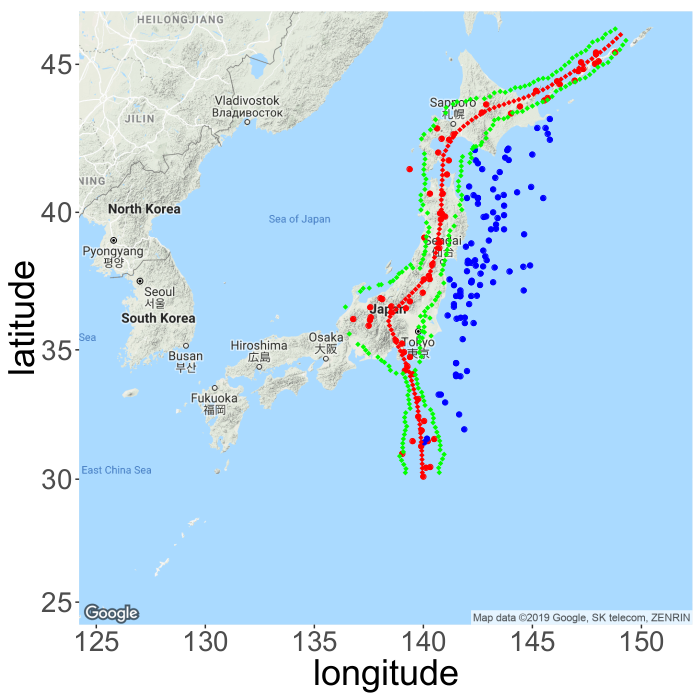}
\caption{}
\end{subfigure}
\hspace{1 in}
\begin{subfigure}[b]{0.3\textwidth}
\includegraphics[width=2 in]{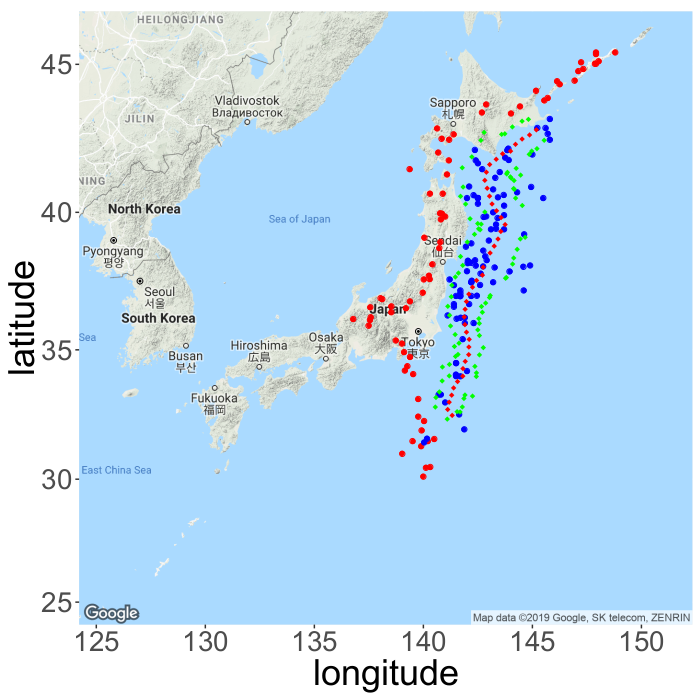}
\caption{}
\end{subfigure}\\
%\hspace{0.1 in}
\begin{subfigure}[b]{0.3\textwidth}
\centering
\includegraphics[width=2 in]{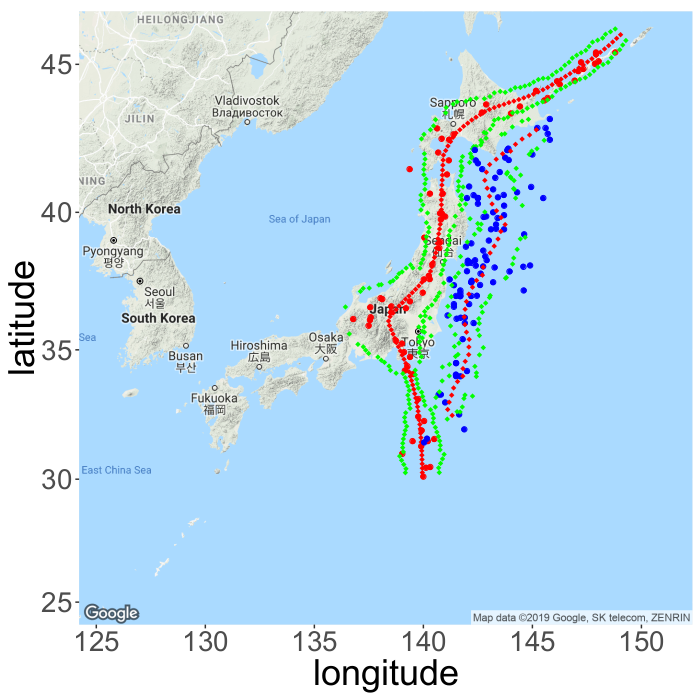}
\caption{}
\end{subfigure}
\caption{Principal flows (in red) with margins (in green) from the offshore earthquakes ((a), in blue) and volcanoes ((b), in red) data with specific starting points and scale parameters $h_{\text{vol}}=0.035$, $h_{\text{eq}}=0.02$. Misclassified offshore earthquake data ((c), in blue within volcanoes flow margins) from the case with scale parameters $h_{\text{vol}}=0.035$, $h_{\text{eq}}=0.02$.}
\label{flow_offshore}
\end{figure}

To highlight the results, we plot the principal boundaries with margins from the case with specific scale parameters $h_{\text{vol}}=0.035$, $h_{\text{eq}}=0.02$ in Figure \ref{flow_offshore}(a)-(b). We report that there are 0 misclassified points for the volcano data and 2 misclassified points for the offshore earthquake data. The misclassified offshore earthquakes are plotted in Figure \ref{flow_offshore}(c). We can see that the principal boundary does a relatively  good job on this data, separating the two data clusters well, except for the two instances where the offshore earthquake data is  mixed with the volcano data in the south.

%%%%%% plot of flow
 
  %%%%%% plot of misclassified data
%\begin{figure}[h]
%\centering
%\includegraphics[width=3 in]{PDF_Data/misclassified_offshore.png}
%\vspace{0.2 in}
%\caption{Plot of misclassified offshore earthquake data (in blue) from the case with scale parameters $h_{\text{vol}}=0.035$, $h_{\text{eq}}=0.02$.}\label{flowmis_offshore}
%\end{figure}

\section{Conclusion} 
The classification problem for data points on non-linear manifolds is a very challenging topic and plays an increasingly important role in real-world problems. %Conventional approaches, such as SVM in the Euclidean space, are essentially unhelpful for either learning the structure of the underlying manifold or deciding the boundary directly on the manifold. 
%The main reason for this lies in the fact that those approaches simply do not use the local geometric information of the manifold. 
With the aim of finding a new method for classifying data points with labels on manifolds, we have proposed a new framework built directly on the manifold. The key techniques we have used involve the tangent space at a given point of the principal flow, which in principle represents the local geometry of the data variation. In other words, the type of local geometric structure we use is the local vector field of the principal flow. We showed the necessity of estimating the margin between the boundary and the flows from a classification perceptive. The combination of fine-tuning the vector fields and the margin provides the alignment of the local data geometry and the global coordinates of the boundary on the manifold. The principal boundary for a finite sample was seen to be interpretable as a local equivalence of the classification boundary by SVM with an analysis. We claim here that although the principal boundary coincides with the SVM locally according to Theorem \ref{thm:equivalence}, in practice they appear to be quite different as the basis (kernel) functions that one would use in SVM are usually unknown. We have shown the convergence of the random principal boundary to its continuous counterpart  following a probability distribution on the manifold. Examples related to the implemented algorithm of the principal boundary and the numerical comparison with that processed by SVM, random forest and $K$-means, with the goal of classification, are demonstrated. % thus providing a nice alternative to the flows.
%This is due to the parameterization of the sub-manifold that we use in this paper.

%Regarding the issue of choosing the locality parameter $h$, or equivalently, at which scale of the local covariance should one consider? Admittedly, different sub-manifolds would have been fitted by choosing different parameters. Still, we suggest not making a strict statement on optimizing the $h$; rather, one should overview a sequence of $h$. We recommend the readers engage in a discussion of such choices associated with possible forms of criterion in \cite{Panaretos2014} and the scale space perspective \cite{Chaudhuri2000}. Simultaneously, we were able to define the principal sub-manifold to any dimension $k \leq d$, and this may also be seen as the development of a heuristic understanding of backward stepwise principle of PCA on manifolds: in the backward PCA, the best approximating affine subspaces are constructed from the highest dimension to the lowest one, see \cite{backward2010} for a spherical case of subspace, while in the case of principal sub-manifold, each net of the principal sub-manifolds (i.e., the principal directions) corresponds to the lower dimension sub-manifolds, compared to the entire sub-manifold.

The formulation of the principal boundary can be extended to several lines of research. From the theoretical point of view, Condition 5.1 (covering ellipse ball) can essentially be relaxed to suit the needs of the analysis for a finite sample. This, if done with a detailed error analysis, would potentially help us understand the boundary better and improve accuracy. One referee pointed out that the principal boundary may be related to a kind of medial manifold between the two principal manifolds.  Although the boundary is conceptually connected to the medial sheet in \cite{medial2008}, the latter is only defined in Euclidean space with a dimension of not more than $3$. Moreover, the medial sheet does not necessarily enjoy the smooth property like the boundary does. From the application point of view, this new method has the potential to be a useful tool for real data analysis for manifolds with dimensions greater than 2. In principle, the generalization for % ($m>2$) is possible. There are two sub-problems involved: the first one is to extend the flows to high dimension, and the second one is to extend the boundary to high dimension. For the former problem, the current setting is extendable \cite{prin-sub2017}; for the latter problem, the sub-manifolds $\gamma_1$ and $\gamma_2$ are essentially co-dimension one sub-manifolds. To find the principal boundary $\gamma_c$ (now it is a sub-manifold with dimension $m-1$) from $\gamma_1$ and $\gamma_2$,
higher dimensional manifolds ($m>2$) is possible by extending both the flows and boundary to high dimension. In practice, the current setting of flows is extendable \cite{prin-sub2017}. Since the sub-manifolds $\gamma_1$ and $\gamma_2$ are essentially co-dimension one sub-manifolds, the extended principal boundary $\gamma_c$ should be also a sub-manifold with dimension $m-1$. Hence, 
one has to learn the tangent space of $\gamma_c$ from the tangent space of $\gamma_1$ and $\gamma_2$. A similar idea applies but the treatment involved will be significantly different from the current one (see Appendix A \cite{Boundarysupp}). Certainly, a successful classification also depends on %1) the data configuration; 2) the noise. 
the data configuration and the noise. 
If the labeled data on the manifold overlaps significantly, one might consider using penalty functions, on top of the adherence used in the current setting. Some of the results in non-parametric regression or machine learning will be helpful in this respect. As this is one of our ongoing works, we will investigate it in the future.

\begin{comment}
\section{Technical Details} 
\subsection{Evaluation of vector field and the derivative of vector field}

Even though the term $M(\gamma)$ is essentially known for $\gamma$ after solving two principal flows $\gamma_1$ and $\gamma_2$, it requires evaluating the derivative of vector field $\dot{W}(\gamma_1)$ (or $\dot{W}(\gamma_2)$) as well as $\dot{\gamma}_1$ (or $\gamma_2$). Here, we give the numerical evaluation of these two terms based on the extension of vector field.

For a neighborhood $N(x_c)$ of $x_c$ in $\mathbb{R}^d$, we extend the tangential vector field to a vector field on $N(x_c)$ such that $W(x):= \left\{W(x), x \in N(x_c) \cap \mathcal{M}\right\}$. By this extension, we know that  $W(x)$ is essentially a function from $N(x_c) \in \mathbb{R}^d$ to $\mathbb{R}^d$. hence, for any $x \in N(x_c)$, the derivative of the vector field $W(x)$ can be approximated by numerical differentiation. In particular,
\begin{equation*}
\dot{W}(x)_{i,j}=\frac{-W(x+\epsilon_i)_j+8W(x+\epsilon_i)_j-8W(x-\epsilon_i)_j+W(x-2\epsilon_i)_j}{12 \epsilon}
\end{equation*} 
where $W(x)_j$ is the $j$th component of $W(x)$ and $\epsilon_i=\epsilon e_i$, where $e_i= (0, \cdots, 1_i, \cdots, 0)$.
\end{comment}

\section*{Acknowledgements}
We are grateful to Professor Steve Marron for intellectual support and to MOE Tier 2 funding (R-155-000-184-112) at the National University of Singapore for financial support. This work was also supported in part by NSFC 11571312 and 91730303.
We thank Zengyan Fan and Wee Chin Tan for reading our manuscript and providing helpful comments on the manuscript.

\newpage

\bibliographystyle{abbrv}
\bibliography{Zhigang}

\end{document}